\documentclass[runningheads]{llncs}
\usepackage[T1]{fontenc}
\usepackage{graphicx}
\usepackage{amsmath,amssymb} % define this before the line numbering.

\usepackage{tikz}
\usepackage[acronym]{glossaries}
\usepackage{comment}
\usepackage{amsmath,amssymb} % define this before the line numbering.
\usepackage{color}
\usepackage{booktabs}
\usepackage{multirow}
\usepackage{color,soul}
\usepackage{ulem}
\usepackage{float}
\usepackage{subcaption}
\usepackage{pdflscape}

\usepackage[capitalize]{cleveref}
\crefname{section}{Sec.}{Secs.}
\Crefname{section}{Section}{Sections}
\Crefname{table}{Table}{Tables}
\crefname{table}{Table}{Tables.}
\begin{document}

\title{Object Detection in Foggy Scenes by Embedding Depth and Reconstruction into Domain Adaptation}
\titlerunning{Detection in Fog by Embedding Depth and Reconstruction into DA}
% If the paper title is too long for the running head, you can set
% an abbreviated paper title here
%
\author{Xin Yang\inst{1}\orcidID{0000-0002-4617-5733} \and
	Michael Bi Mi\inst{2} \and
	Yuan Yuan\inst{2}  \and
	Xin Wang\inst{2}  \and
	Robby T. Tan\inst{1,3}\orcidID{0000-0001-7532-6919}}
\authorrunning{X. Yang et al.}
% First names are abbreviated in the running head.
% If there are more than two authors, 'et al.' is used.
%
\institute{National University of Singapore,
	Huawei International Pte Ltd,Yale-NUS College \\
	\email{e0674612@u.nus.edu},
	\email{michaelbimi@yahoo.com},
	\email{\{yuanyuan10,wangxin237\}@huawei.com},
	\email{robby.tan@\{nus,yale-nus\}.edu.sg}}
\maketitle              % typeset the header of the contribution

\begin{abstract}
Most existing domain adaptation (DA) methods align the features based on the domain feature distributions and ignore aspects related to fog, background and target objects, rendering suboptimal performance. 
In our DA framework, we retain the depth and background information during the domain feature alignment. 
A consistency loss between the generated depth and fog transmission map is introduced to strengthen the retention of the depth information in the aligned features.
To address false object features potentially generated during the DA process, we propose an encoder-decoder framework to reconstruct the fog-free background image. 
This reconstruction loss also reinforces the encoder, i.e., our DA backbone, to minimize false object features.
Moreover, we involve our target data in training both our DA module and  our detection module in a  semi-supervised manner,  so that our detection module is also exposed to the unlabeled target data,  the type of data used in the testing stage.
Using these ideas, our method  significantly outperforms the state-of-the-art method (47.6 mAP against the 44.3 mAP on the Foggy Cityscapes dataset), and obtains the best performance on multiple real-image public datasets.
Code is available at: https://github.com/VIML-CVDL/Object-Detection-in-Foggy-Scenes

\end{abstract}
\section{Introduction}
\label{sec:intro}
Object detection is impaired by bad weather conditions, particularly fog or haze. 
Addressing this problem is important, since many computer vision applications, such as self-driving cars and video surveillance, rely on robust object detection regardless of the weather conditions.
One possible solution is to employ a pre-processing method, such as image defogging or dehazing \cite{zhang2018densely,liu2019griddehazenet} right before an object detection module. 
However, this solution is suboptimal, since bad weather image enhancement itself is still an open problem, and thus  introduces a risk of removing or altering some target object information.

\begin{figure}[t]
	\centering
	\begin{subfigure}[b]{.495\textwidth}
		\centering
		% include first image
		\includegraphics[width=\linewidth, height=3.1cm]{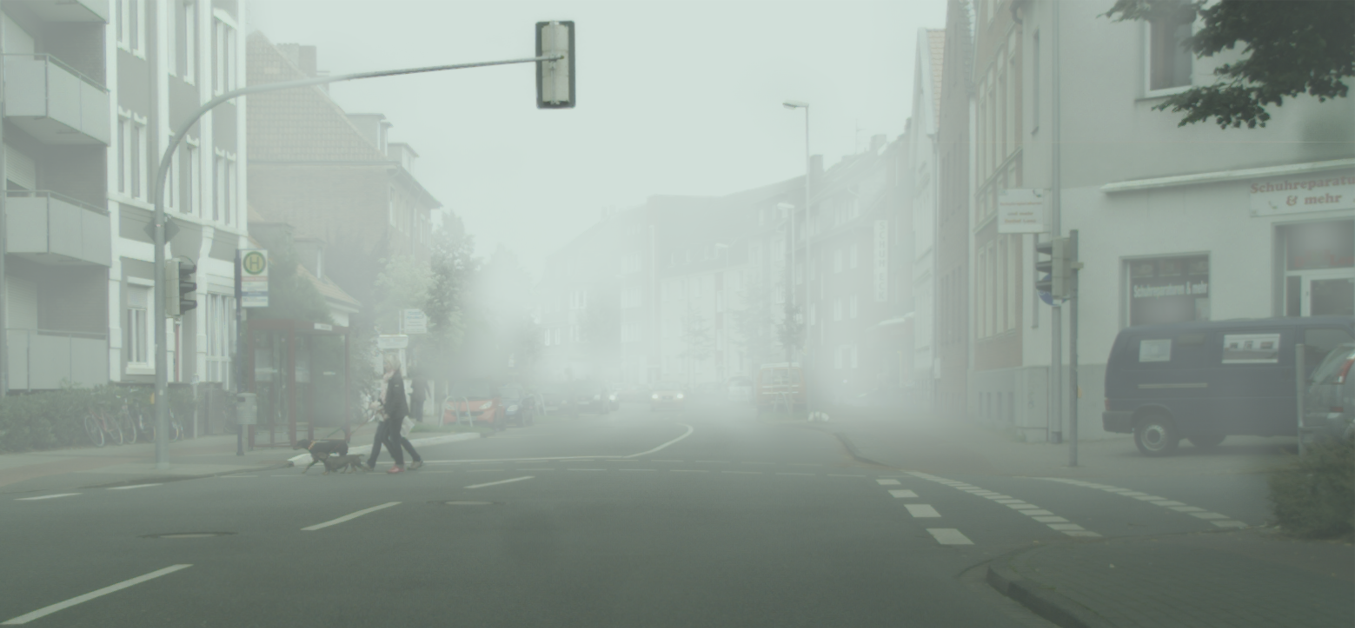}  
		\caption{Input Image}
		\label{fig:sub-first}
	\end{subfigure}
	\begin{subfigure}[b]{.495\textwidth}
		\centering
		% include second image
		\includegraphics[width=\linewidth, height=3.1cm]{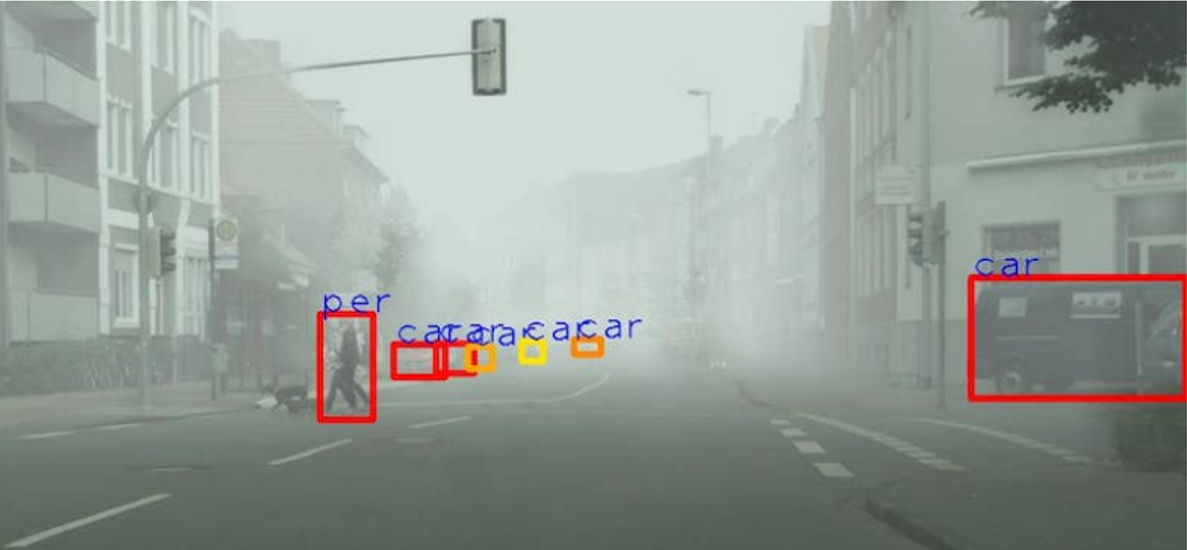}  
		\caption{PBDA\cite{sindagi2020prior}}
		\label{fig:sub-second}
	\end{subfigure}
	\\
	\begin{subfigure}[b]{.495\textwidth}
		\centering
		% include third image
		\includegraphics[width=\linewidth, height=3.1cm]{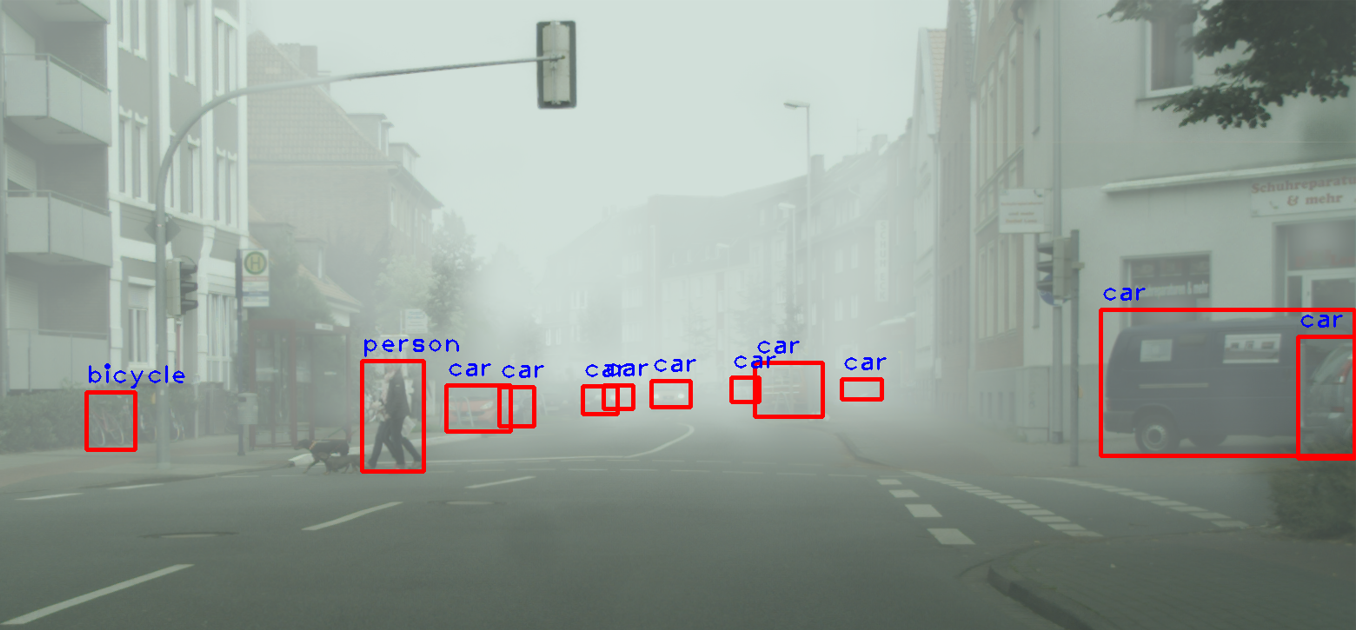}  
		\caption{\textbf{Our Result}}
		\label{fig:sub-third}
	\end{subfigure}
	\begin{subfigure}[b]{.495\textwidth}
		\centering
		% include third image
		\includegraphics[width=\linewidth, height=3.1cm]{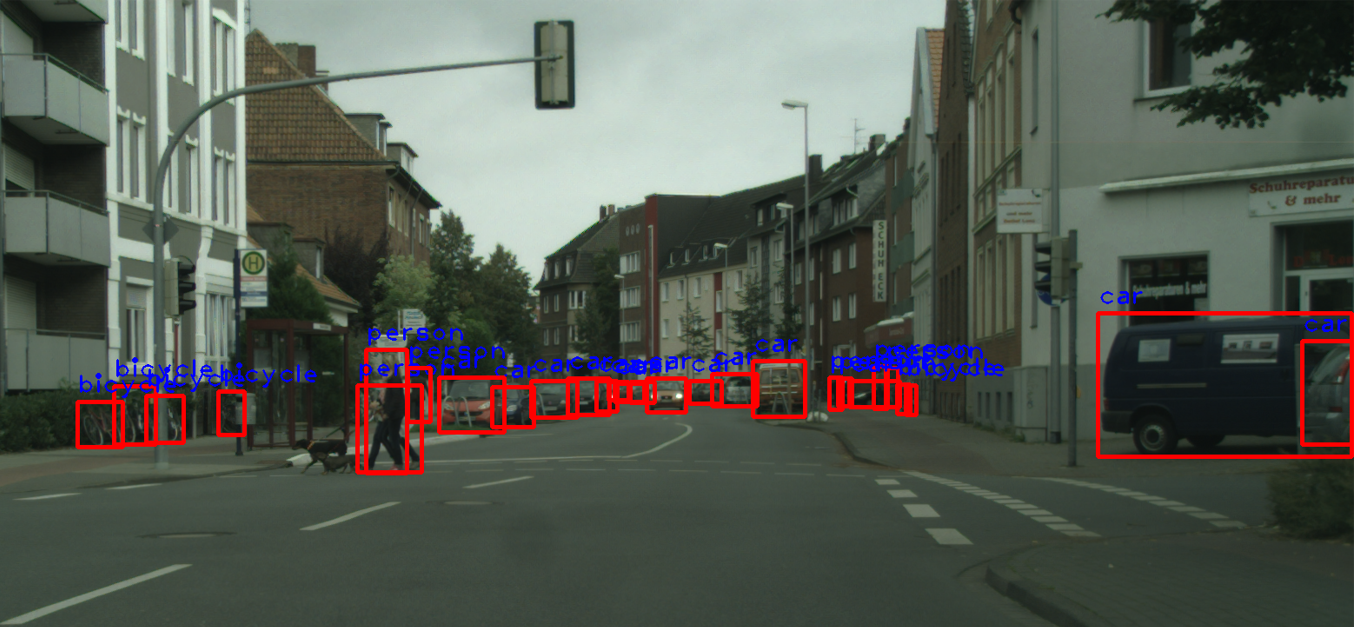}  
		\caption{Ground-Truth}
		\label{fig:sub-third}
	\end{subfigure}
	\caption{\label{fig:results} (a) Input dense fog image from Foggy Cityscapes~\cite{cordts2016cityscapes}. (b) Result from PBDA \cite{sindagi2020prior}, where many objects are undetected. (c) Ours, where more objects are detected. (d) Ground-truth, annotated from its corresponding clear image. Zoom in for better visualization.}
\end{figure}

Recently, object detection methods based on unsupervised domain adaptation (DA) (e.g.,~\cite{ganin2015unsupervised,chen2018domain,chen2021scale}) have shown promising performance for bad weather conditions.
By aligning the source (clear weather image) and the target (weather degraded image) distributions in the feature level, a domain adaptive network is expected to produce weather-invariant features.
Unlike image pre-processing methods, domain adaptive detection networks do not require an additional defogging module during the inference stage, and can also work on both clear and foggy conditions.
The DA methods, however,  were not initially proposed for the adverse weather conditions, and hence align the source and target features based only on feature alignment losses, ignoring some important aspects of the target data, such as depth, transmission map, reconstruction of object instances of the target data, etc.
This is despite the fact that for bad weather, particularly fog or haze, these aspects can be imposed.
Sindagi et al.~\cite{sindagi2020prior} attempt to fuse DA with adverse weather physics models, but do not obtain a satisfactory performance (39.3 mAP on Foggy Cityscapes, compared to 47.6 mAP of ours).

In this paper, we propose a DA method that learns domain invariant features by considering depth cues, consistency between depth from fog transmission, and clear-image reconstruction.
Moreover, we involve the source and target data to train our whole network in a semi-supervised manner, so that our object detection module can be exposed to the unlabeled target data, which has the same type as data in the testing stage.
Most existing DA methods aim to suppress any discrepancies between the source and target data in the feature space, and 
this includes any depth distribution discrepancies.
However, depth information is critical in object detection \cite{ding2020learning,vu2019dada,saha2021learning}, and thus the depth suppression will significantly affect the object detection performance.
To resolve this, we propose a depth estimation module and its corresponding depth loss, so that  the depth information in our features can be retained.
This depth loss forces our DA backbone to retain the depth information during the DA process.
Moreover, to further reinforce our DA backbone to retain the depth information, we also add a transmission-depth consistency loss. 

When performing DA, the source and target images are likely to contain different object instances, and aligning two different objects with different appearances encourages the generation of false features. 
To address this problem, we fuse a reconstruction module into our DA backbone, and propose a reconstruction loss.
Based on the features from our DA backbone's layers from the target (fog image), our reconstruction module generates a clear image. 
Our reconstruction loss thus measures the difference between the estimated clear image and the clear-image pseudo ground-truth of the target obtained from an existing defogging method.
This reconstruction loss will then prevent false features generated during the feature extraction process.
During the training stage, the DA model gradually becomes more robust to fog, and the predictions on the unlabeled target images  become more reliable.
This gives us an opportunity to employ the target data to train our object detection module, so that the module can be exposed to both source and target data and becomes less biased to the source data.

\cref{fig:results} shows our object detection result,  which  incorporates all our losses and ideas into our DA backbone.
As a summary, our contributions and novelties are as follow:
\begin{itemize}
	\item 	Without imposing our depth losses, DA features are deprived from the depth information, due to the over-emphasis on source/target adaptation. This deprivation negatively affects object detection performance. Hence, we introduce  depth losses to our  DA backbone to retain the depth information in our DA features.
	\item We propose to reinforce the target transmission map to have consistent depth information to its corresponding  depth estimation. 
	This consistency loss constraints the transmission map and improves further the DA performance and the depth retention.
	\item We propose to integrate an image reconstruction module into our DA backbone. 
	Hence, any additional false object features existing in DA features will be penalized, and hence minimized.
\end{itemize}
Our quantitative evaluations show that our method outperforms the state-of-the-art methods on various public datasets, including real image datasets.

\section{Related Work}
\label{sec:relatedwork}

\noindent{\bf Object Detection in Foggy Scenes}
Most existing object detection models require a fully-supervised training strategy \cite{zou2019object}.
However, under adverse weather conditions, having sufficient images and precise annotations is intractable.
A possible solution is to utilize defogging algorithms.
The defogged images are less affected by fog, and hence they can be fed into object detection models which are trained with clear images directly.
However, defogging is still an existing research problem and thus limits the performance potential of this approach.
Moreover, defogging introduces an additional computational overhead,  hindering the real-time process for some applications.
These drawbacks were discussed and analyzed in \cite{sakaridis2018semantic,sindagi2020prior,zhou2021self}.

\noindent{\bf Domain Adaptation}
DA methods were proposed to train a single network which can work on different domains.
In DA, there will be labeled images from the source domain and unlabeled images from the the target domain.
During the training stage, images from both domains will be fed into the network.
The source images with annotations will be used for the training of object detection part.
Meanwhile, a domain discriminator will examine which domain the extracted feature maps come from.
The discriminator will be rewarded for accurate domain prediction, but the network will be penalized.
Hence, the network is encouraged to extract domain invariant feature maps, i.e., fog free features.
Since the feature maps are already domain invariant, the detection trained with the source labels can also detect objects from the target images.
Additionally, once the network is trained, the domain discriminator is not needed anymore, hence no additional overhead in the inference stage.

There are a few existing DA methods that tried to perform DA between clear weather and fog weather.
Some methods investigated where and how to put the domain discriminators, so that the DA can be more efficient whilst retain most object-related information \cite{chen2018domain,saito2019strong,shen2019scl,chen2021scale,zhou2021self}.
\cite{sindagi2020prior,nguyen2020domain,guan2021uncertainty,vs2021mega,huang2022category,zhou2022multi} aimed to designed a more suitable domain discriminator, using transmission maps, entropy, uncertainty masks, memory banks/dictionaries and class clusters.
However, most of these methods focus on synthetic datasets, and ignores the fact where the weather-specific knowledge prior can also be integrated to better describe the domain discrepancy.
%

%\subsection{?Semi-supervised Learning}

\begin{figure*}[t]
	\centering
	\includegraphics[width=\textwidth]{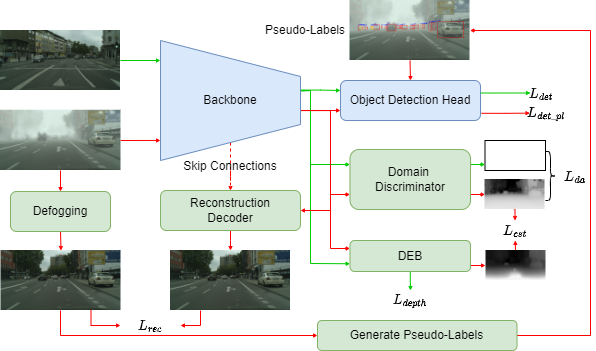}
	\caption{\label{fig:architecture}The network consists of five parts. (1) Backbone extracts feature maps from the input images. (2) Object detection head localizes and categorizes object instances from the feature maps. (3) Domain discriminator and DEB encourage the backbone to extract fog-invariant features, and maintains the images' depth distributions. (4) Reconstruction decoder minimizes the fake object features generated by DA. (5) Pseudo-Labels involve target domain information in the pipeline, and apply consistency regularization between fog and defogged images. The green arrows represent source data-flow, and the red arrows represent target data-flow. Note, only the blue modules are needed in the testing stage.}
	
\end{figure*}

\section{Proposed Method}
\label{sec:proposed method}

\cref{fig:architecture} shows the pipeline of our method, where clear images are our source input, and  foggy images are our target input.
For the source images, we have their corresponding annotations (bounding boxes and classes) to train our object detection module.
For the target images, we do not have any annotations.
In this DA framework, we introduce a few constraints: depth, consistency between the transmission and depth, and clear background reconstruction. 
The goal of adding these constraints is to extract features from both source (clear image) and target (fog image) that are robust for object detection.
Moreover, we exploit the target predictions to train our object detection module, so that the module can be exposed to the unlabeled target data, and hence less biased to the source data.

\noindent\textbf{Object Detection}
For the object detection module, we employ Faster-RCNN~\cite{ren2015faster}, which consists of  a backbone $\mathcal{F}$  for feature extraction, and an object detection head, $\mathcal{G}$.
The loss for the object detection is defined as:
\begin{equation}
	\mathcal{L}_{\text{det}}=\mathcal{L}_{\text{rpn}}+\mathcal{L}_{\text{cls}}+\mathcal{L}_{\text{bbox}},
	\label{eq:objectdetection}
\end{equation}
where, $\mathcal{L}_{\text{rpn}}$ is the regional proposal loss, $\mathcal{L}_{\text{cls}}$ is the classification loss, and  $\mathcal{L}_{\text{bbox}}$ is the localization loss.

\noindent\textbf{Domain Adaptation}
Our domain adaptation backbone shares the same backbone as that of  the object detection module.
For the domain discriminator, we use transmission maps as the domain indicator (i.e., the discriminator is expected to produce a blank map for source, and a transmission map for target). 
The corresponding loss can be defined as:
\begin{equation}
	\label{eq:daloss}
	\begin{aligned}
		\mathcal{L}_{\text{da}}=\left\|\mathcal{D}(\mathcal{F}(I_s))\right\|^2_2
		+\left\|t-\mathcal{D}(\mathcal{F}(I_t))\right\|^2_2,
	\end{aligned}
\end{equation}
where, $\mathcal{F}$ is the backbone, $\mathcal{D}$ is the domain discriminator.
$I_s$ and $I_t$ are the input images from the source domain and the target domain, respectively.
$t$ is the transmission map for the target image.

\subsection{Depth Estimation Block (DEB)}
In the DA process, it is unlikely that a pair of source and target images to have the same depth distribution, since they unlikely contain the same scenes.
Thus, when the existing DA methods suppress the domain discrepancies, the depth information is also suppressed in the process.
However, recent methods have shown that the depth information benefits object detection~\cite{ding2020learning,vu2019dada,saha2021learning}, which implies that suppression of depth can affect the performance of object detection.

\begin{figure}[t]
	\centering
	\begin{subfigure}{0.495\linewidth}
		\includegraphics[width=1\linewidth]{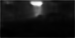}
		\caption{Depth estimation  1}
	\end{subfigure}
	\begin{subfigure}{0.495\linewidth}
		\includegraphics[width=1\linewidth]{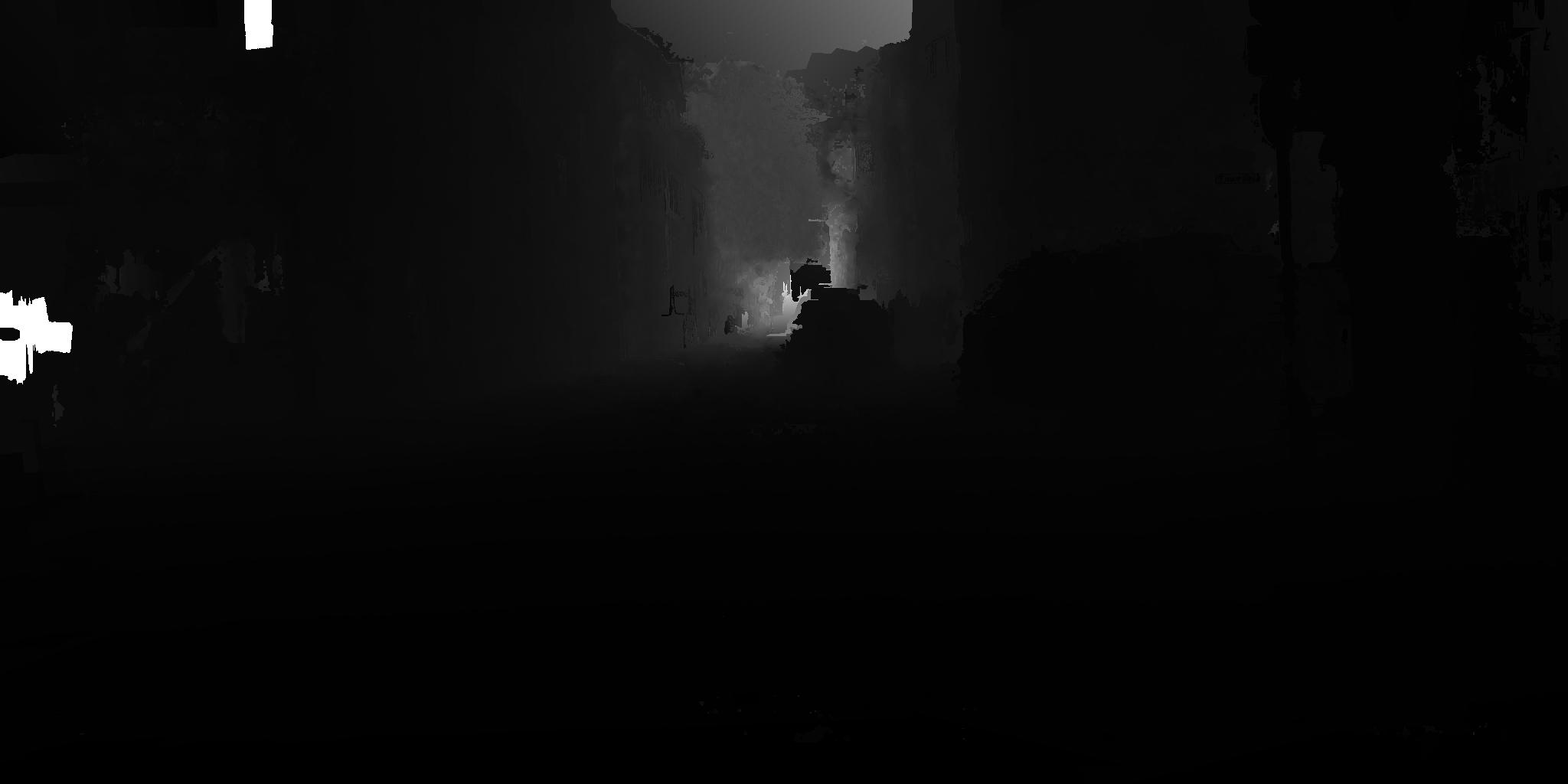}
		\caption{Depth ground-truth 1}
	\end{subfigure}
	\\
	\centering
	\begin{subfigure}{0.495\linewidth}
		\includegraphics[width=1\linewidth]{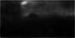}
		\caption{Depth estimation  2}
	\end{subfigure}
	\begin{subfigure}{0.495\linewidth}
		\includegraphics[width=1\linewidth]{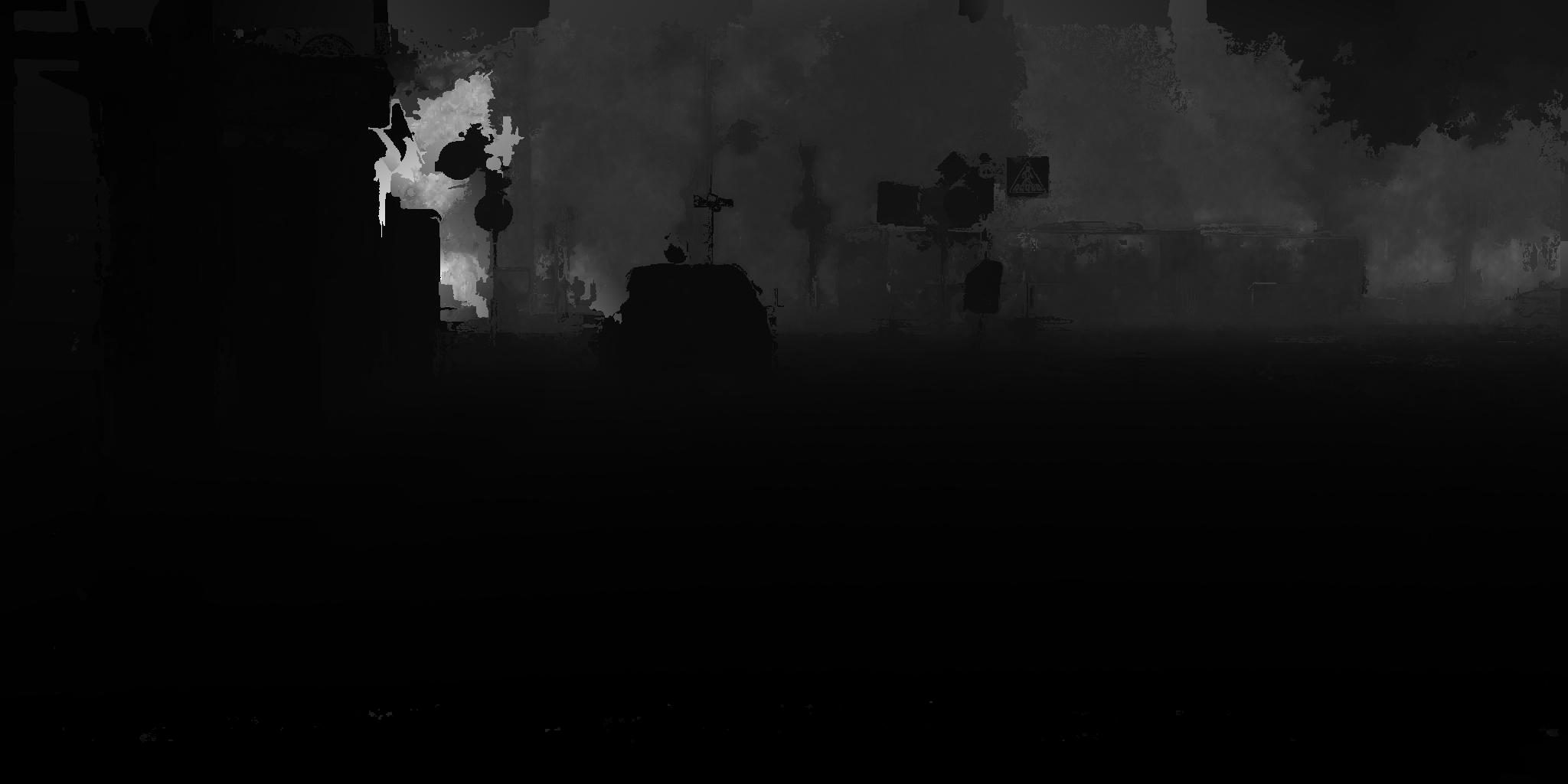}
		\caption{Depth ground-truth 2}
	\end{subfigure}
	\caption{\label{fig:dep} Examples of the estimated depths in comparison with the ground-truths. Note that, the estimated depths look blurry as they have a low resolution ($38\times75$). We can observe that the estimated depth maps (a) and (c) from DEB match the depth distribution patterns in the ground-truth depth maps (b) and (d), respectively, which indicates that our feature maps retain the depth information.}
	\label{fig:depth}
\end{figure}

To address this problem, we need to retain the depth information during the DA feature alignment.
We introduce DEB, a block that generates a depth map based on the extracted features from our DA backbone.
We define the depth recovery loss as follows:
\begin{equation}\label{eq:deb}
	\mathcal{L}_{\text{depth}}=\left\|\text{DEB}(\mathcal{F}(I_s))-D_{gt}\right\|^2_2,
\end{equation}
where, $D_{gt}$ is the ground-truth depth map, which is resized to the same size as the corresponding feature map.
$\mathcal{F}(I_s)$ represents the source feature maps, and $\text{DEB}()$ is our DEB module.
For datasets such as Cityscapes \cite{cordts2016cityscapes}, they provide the ground-truth depth maps. 
For the other datasets which do not provide depth ground-truth, we need to generate the depth maps as a pseudo ground-truth using the existing depth estimation networks, such as \cite{godard2017unsupervised,chen2019towards,godard2019digging}.
Note, DEB is only trained on the source images.
Unlike the transmission DA loss in Eq.~(\ref{eq:daloss}), we backpropagate the depth loss over both our DA backbone and DEB to retain the depth information in our DA features.
Fig.~\ref{fig:depth} shows our depth estimations. Note that our goal here is not to have accurate depth estimation, but to retain depth cues in our features.

\subsection{Transmission-Depth Consistency}
In foggy scenes, we can model the transmission of light throughout the fog particles as $t =\exp(-\beta D)$, where $t$ is the transmission, $D$ is the depth, and $\beta$ as the fog particles attenuation factor. As one can notice, there is a strong correlation between the transmission and depth. 
Hence, we reinforce our predicted transmission and depth to be consistent:
\begin{equation}\label{eq:debcst}
	\begin{aligned}
		\mathcal{L}_{\text{cst}}=\left\|Norm(-\log(\mathcal{D}	(\mathcal{F}(I_t))))-Norm(\text{DEB}(\mathcal{F}(I_t)))\right\|^2_2,
	\end{aligned}
\end{equation}
where $\mathcal{D}(\mathcal{F}(I_t))$ is the generated transmission map from the domain discriminator, and $\text{DEB}(\mathcal{F}(I_t))$ represents the estimated depth map. 
$Norm()$ represents a normalization operator.
Like most defogging methods, we assume that $\beta$ is uniform across the input target.
Since the transmission and the depth values are the same only up to a scale, we normalize their values, and thus consequently cancel out $\beta$ in the process.
This consistency loss enforces the consistency between the depth encoded in the estimated transmission and the depth from our DEB, this constraint leads to more robust depth information in our features.

\subsection{Reconstruction DA}

Since DA methods use unpaired source and target that most likely contain different  object instances, when the DA backbone aligns the features, the alignment will occur on two different object instances (e.g., car in the source, and motorbike in the target). Hence, when the DA backbone is suppressing such discrepancy, consequently it can generate false features, which can harm the object detection performance.

To address this problem, we regularize the generation of object features by fusing a reconstruction decoder into our DA backbone.
This decoder reconstructs the features back to the clear background image.
To train this decoder, we use either the clear image or the defogged image of the target image as the ground-truths.
Since the reconstruction ground-truths have the same object instances as in the target images,  the reconstruction loss will prevent our DA backbone from generating false instance features. Our reconstruction loss  is defined as:
\begin{equation}\label{eq:rec}
	\mathcal{L}_{\text{rec}}=\left\|\mathcal{R}(\mathcal{F}(I_t))-I^{de}\right\|^2_2,
\end{equation}
where $\mathcal{R}$ is the reconstruction module and $\mathcal{R}(\mathcal{F}(I_t))$ represents the reconstructed target image.
$I^{de}$ is the clear/defogged target image, which we use as the ground-truth for the reconstruction.
Only target images are involved in the reconstruction, as there are no fog distortions in the source images.

\subsection{Learning from Target}

In many DA cases, target data does not have ground-truths, and hence \cite{sindagi2020prior,chen2018domain,chen2021scale,zhou2021self,saito2019strong,shen2019scl,nguyen2020domain,guan2021uncertainty,vs2021mega} use the source domain's knowledge to train the object detection module, right after the domain feature adaptation process.
This means the object detection module is never exposed to the target data, which is likely to be different from the source data in terms of the appearances of the object instances (e.g., the shapes of road signs in one country are different from those of the road signs from another country, etc.).
%
%Moreover, involving the target data in training the object detection module is important, since it means we have more data to training the module, and make our object detection module less biased to the source data. 
%
The main problem why many methods do not use target data in training the detection module is because the target data is unlabeled.

With the help from our DA module, our detection module's performance on the target data is improving over iterations.
As the predictions on target become more accurate, we can select some reliable predictions as pseudo-labels to train our detection module.
Hence, our training is split into two stages for each iteration: 
In the first stage, we generate pseudo-labels from the whole network; 
in the second stage, we do DA training involving the generated pseudo-labels.
To obtain more reliable predictions in the first stage, we employ a defogging method to augment our target input.
The augmented target input in the form of defogged image will enable our network to estimate the bounding boxes and class labels. 
If the network has high confidence with these estimates, it means the estimates can be considered as reliable and used as pseudo-labels.
In the second stage, we feed  the same target input to our network without augmentation, and let it to predict the bounding boxes and class labels.
We then enforce  these network's estimates to be consistent with the pseudo-labels.
This process encourages the network to become more robust to fog, and to expose our object detection module to the target data.
Note that, in the end of each iteration, we employ Exponential Moving Average (EMA) in our network to generate more reliable predictions.

\noindent\textbf{Total Loss} Combining all the losses we introduced above, we can derive our overall loss as:
\begin{equation}\label{eq:overall}
	\mathcal{L}=\mathcal{L}_{\text{det}}+\lambda\mathcal{L}_{\text{da}}+a\mathcal{L}_{\text{depth}}+b\mathcal{L}_{\text{cst}}+c\mathcal{L}_{\text{rec}} + \mathcal{L}_{\text{det\_pl}},
\end{equation}
where, $\lambda, a, b, c$ are the weight parameters to control the importance of the losses. 
$\mathcal{L}_{\text{det\_pl}}$ is the detection loss with pseudo-labels.
Note that, in the testing stage, we only use our DA backbone and the detection module. 
In other words, all the additional modules (i.e., the domain discriminator, the depth estimation block, the reconstruction module or the pseudo-labels) does not affect the runtime in the testing stage.

%------------------------------------------------------------------------- 

\section{Experimental Results}
\label{sec:results}
We compare our DA method with recent DA methods:  \cite{sindagi2020prior,chen2018domain,chen2021scale,zhou2021self,saito2019strong,shen2019scl,nguyen2020domain,guan2021uncertainty,vs2021mega,huang2022category,zhou2022multi}, where the last two are published as recent as this year.
To make the comparison fair, we use the same base of object detection, which is Faster-RCNN \cite{jjfaster2rcnn}.
For the backbone, our method uses a pretrained ResNet-101~\cite{jjfaster2rcnn}.
We set the confidence threshold $\tau$ to be 0.8 for all the experiments.
More details of our blocks can be found in the supplementary material.
Our overall network is trained end-to-end.
We follow the same training settings as in \cite{sindagi2020prior,chen2018domain,chen2021scale,saito2019strong,shen2019scl}, where the networks are trained for 60K iterations, with a learning rate of 0.002.
We decrease the learning rate by a factor of 10 for every 20K iterations.
The weights parameters $\lambda, a, b, c$ are empirically set to be $0.1, 10, 1, 1$, respectively.

As for the datasets, the Cityscapes dataset is a real world street scene dataset provided by \cite{cordts2016cityscapes}, and all images were taken under clear weather.
Based on this dataset, \cite{sakaridis2018semantic} simulates synthetic fog on each clear image, and creates the Foggy Cityscapes dataset.
We use the same DA settings as in \cite{sindagi2020prior,chen2018domain,chen2021scale,saito2019strong,shen2019scl}, where 2975 clear images and 2975 foggy images are used for training, and 495 foggy images are used for evaluation.

Aside from the Cityscapes dataset, STF (Seeing Through Fog) \cite{bijelic2020seeing}, FoggyDriving \cite{sakaridis2018semantic}, and RTTS \cite{li2018benchmarking} are the datasets with real world foggy images used in our experiments.
STF dataset categories its images into different weather conditions, we choose \textit{clear weather daytime} as our source domain and \textit{fog daytime} as our target domain.
We randomly select 100 images from \textit{fog daytime} as our evaluation set, and use the rest to train the network.
For RTTS, we follow the same DA settings as in \cite{sindagi2020prior,shen2019scl}.
For FoggyDriving, it only contains 101 fog images, which is insufficient for DA training.
Hence, we evaluate the DA models trained on Cityscapes/Foggy Cityscapes directly on these datasets. 

\subsection{Quantitative Results}

The synthetic Foggy Cityscapes dataset has the ground-truth transmission maps, depth maps and the clear background of the target images for reconstruction, thus we can use the ground-truths directly in our training process, however for fair comparisons we do not use them. Instead we employ DCP \cite{he2010single} to compute the transmission maps, reconstruction maps, and use it as the defogging pre-processing module when involving target predictions. 
As for the depth, we employ Monodepth \cite{godard2017unsupervised} to compute the pseudo ground-truths.
We also employ DCP and Monodepth  for real data that have no ground-truths of clear images, transmission maps, and depth maps.
Note that, the methods we use to generate pseudo ground-truths  (DCP and Monodepth) are not the state-of-the-art methods, as we want to show that our DA's performances are not limited by the precision of the pseudo ground-truths.

The results on this dataset are provided in \cref{tab:result}.
The mAP threshold for all the models is 0.5.
%.
When comparing DA models, there are two important non-DA baseline models that need to be considered.
One is the model trained on clear images but tested on foggy images, which we call Lowerbound.
Any DA models should performance better than this Lowerbound model.
In our experiment, Lowerbound is 28.12 mAP for Foggy Cityscapes dataset.
The other model is both trained and tested on clear images, which we call Upperbound.
Since it is not affected by fog at all, the goal of DA models is to approach its performance, but it is not possible to exceed it.
In our experiment, Upperbound is 50.08 mAP for Foggy Cityscapes dataset.
\cref{tab:result} shows that our proposed method performs better than any other DA methods.

For the real world datasets, we cannot compute Lowerbound's and Upperbound's performance, since we do not have the clear background of the foggy images.
Thus, we can only compare our models with the performance of the other DA methods.
The results are presented in \cref{tab:stfresult,tab:foggyresult,tab:rttsresult}.
Our model achieved a better performance on all the real world datasets.
For FoggyDriving, we can observe that our model trained on synthetic dataset can also generalize well on the real world image datasets.
Note that, the compared methods are not as many as the previous table, since some methods only provided their performances on the synthetic datasets, and we do not have their data or code to evaluate on the real world datasets.

\begin{table*}[t]
	\centering
	\caption{\label{tab:result}Quantitative results of Ours compared to the existing DA methods evaluated against Foggy Cityscapes testing set. AP (\%) of each category and the mAP (\%) of all the classes. Bold numbers are the best scores, and underlined numbers are the second best scores. Our mAP outperforms the best existing method over 3\%.}s
	\resizebox{1\columnwidth}{!}{
		\begin{tabular}{|cc|c|c|c|c|c|c|c|c|c|c|}
			\hline
			\multicolumn{2}{|c|}{Method}                                         & Backbone      & person & rider & car  & truck & bus  & train & motor & bicycle & mAP   \\ \hline
			\multicolumn{1}{|c|}{Baseline}                    & Faster-RCNN            & ResNet-101    & 32.0   & 39.5  & 36.2 & 19.4  & 32.1 & 9.4   & 23.3  & 33.2    & 28.1 \\ \hline
			\multicolumn{1}{|c|}{\multirow{8}{*}{DA Methods}} & DA-Faster\cite{chen2018domain}        & ResNet-101    & 37.2   & 46.8  & 49.9 & 28.2  & 42.3 & 30.9  & 32.8  & 40.0    & 38.5  \\ \cline{2-12} 
			\multicolumn{1}{|c|}{}                            & SWDA\cite{saito2019strong}             & VGG-16        & 29.9   & 42.3  & 43.5 & 24.5  & 36.2 & 32.6  & 30.0  & 35.3    & 34.3  \\ \cline{2-12} 
			\multicolumn{1}{|c|}{}                            & SCL\cite{shen2019scl}              & ResNet-101    & 30.7   & 44.1  & 44.3 & 30.0  & 47.9 & 42.9  & 29.6  & 33.7    & 37.9  \\ \cline{2-12} 
			\multicolumn{1}{|c|}{}                            & PBDA\cite{sindagi2020prior}             & ResNet-152    & 34.9   & 46.4  & 51.4 & 29.2  & 46.3 & 43.2  & 31.7  & 37.0    & 40.0  \\ \cline{2-12} 
			\multicolumn{1}{|c|}{}                            & MEAA\cite{nguyen2020domain}             & ResNet-101    & 34.2   & 48.9  & 52.4 & 30.3  & 42.7 & 46.0  & 33.2  & 36.2    & 40.5  \\ \cline{2-12} 
			\multicolumn{1}{|c|}{}                            & UaDAN\cite{guan2021uncertainty}            & ResNet-50     & 36.5   & 46.1  & 53.6 & 28.9  & 49.4 & 42.7  & 32.3  & 38.9    & 41.1  \\ \cline{2-12} 
			\multicolumn{1}{|c|}{}                            & Mega-CDA\cite{vs2021mega}         & VGG-16        & 37.7   & 49.0  & 52.4 & 25.4  & 49.2 & \underline{46.9}  & 34.5  & 39.0    & 41.8  \\ \cline{2-12} 
			\multicolumn{1}{|c|}{}                            & SADA\cite{chen2021scale}             & ResNet-50-FPN & \textbf{48.5}   & \textbf{52.6}  & \textbf{62.1} & 29.5  & 50.3 & 31.5  & 32.4  & \textbf{45.4}    & 44.0  \\\cline{2-12}   
			\multicolumn{1}{|c|}{}                            & CaCo\cite{huang2022category}             & VGG-16 & 38.3   & 46.7  & 48.1 & \underline{33.2}  & 45.9 & 37.6  & 31.0  & 33.0    & 39.2  \\\cline{2-12}  
			\multicolumn{1}{|c|}{}                            & MGA\cite{zhou2022multi}             & VGG-16 & \underline{43.9}   & 49.6  & \underline{60.6} & 29.6  & \underline{50.7} & 39.0  & \underline{38.3}  & 42.8    & \underline{44.3}  \\\cline{2-12}  
			\multicolumn{1}{|c|}{}    & \textbf{Ours}  & ResNet-101    & 39.9   & \underline{51.6}  & 59.0 & \textbf{39.7}  & \textbf{58.0} & \textbf{49.1}  & \textbf{39.2}  & \underline{45.1}    & \textbf{47.6} \\ \hline
		\end{tabular}
	}
\end{table*}

\begin{table}[t]
	\centering
	\caption{\label{tab:stfresult}Quantitative results of Ours compared to the existing DA methods evaluated against the STF testing set. AP (\%) of each category and the mAP (\%).}
	\resizebox{1\textwidth}{!}{\begin{tabular}{|cc|c|c|c|c|}
			\hline
			\multicolumn{2}{|c|}{Method}                                    & PassengerCar   & LargeVehicle   & RidableVehicle & mAP            \\ \hline
			\multicolumn{1}{|c|}{Baseline}                    & Faster-RCNN &          74.0      &       \underline{54.1}        &   \underline{25.7}             &       51.3         \\ \hline
			\multicolumn{1}{|c|}{\multirow{5}{*}{DA Methods}} & DA-Faster   & 77.9          & 51.5          & 25.2    & \underline{51.6}          \\ \cline{2-6} 
			\multicolumn{1}{|c|}{}                            & SWDA        & 77.2          & 50.1          & 24.5          & 50.6          \\ \cline{2-6} 
			\multicolumn{1}{|c|}{}                            & SCL         & 78.1          & 52.5    & 20.7          & 50.4         \\ \cline{2-6} 
			\multicolumn{1}{|c|}{}                            & SADA        & \textbf{78.5} & 52.2          & 24.2          & \underline{51.6}    \\ \cline{2-6} 
			\multicolumn{1}{|c|}{}                            & \textbf{Ours}         & \textbf{78.5} & \textbf{57.4} & \textbf{30.2} & \textbf{55.4} \\ \hline
	\end{tabular}}
\end{table}

\begin{table}[t]
	\centering
	\caption{\label{tab:foggyresult}Quantitative results of Ours compared to the existing DA methods evaluated against the FoggyDriving testing set. mAP (\%) of all the categories.}
	\resizebox{0.42\textwidth}{!}{\begin{tabular}{|cc|c|}
			\hline
			\multicolumn{2}{|c|}{Method}                                  & mAP   \\ \hline
			\multicolumn{1}{|c|}{Baseline}                    & FRCNN     & 26.41 \\ \hline
			\multicolumn{1}{|c|}{\multirow{4}{*}{DA Methods}} & DA-Faster & 31.60 \\ \cline{2-3} 
			\multicolumn{1}{|c|}{}                            & SWDA      & \underline{33.54} \\ \cline{2-3} 
			\multicolumn{1}{|c|}{}                            & SCL       & 33.49 \\ \cline{2-3} 
			\multicolumn{1}{|c|}{}                            & SADA       & 32.26 \\ \cline{2-3} 
			\multicolumn{1}{|c|}{}                            & \textbf{Ours}       & \textbf{34.62} \\ \hline
	\end{tabular}}
\end{table}

\begin{table}
	\caption{\label{tab:rttsresult}Quantitative results of Ours compared to the existing DA methods evaluated against the RTTS testing set. AP (\%) of each category and the mAP (\%).}
	\centering
	\resizebox{0.9\textwidth}{!}{\begin{tabular}{|cc|l|l|l|l|l|l|}
			\hline
			\multicolumn{2}{|c|}{Method}                                                         & person & car   & bus   & motor & bicycle & \multicolumn{1}{c|}{mAP} \\ \hline
			\multicolumn{1}{|c|}{Baseline}                    & Faster-RCNN                            & \underline{46.6}  & 39.8 & 11.7 & 19.0 & 37.0   & 30.9                    \\ \hline
			\multicolumn{1}{|c|}{\multirow{5}{*}{DA Methods}} & DA-Faster                        & 42.5  & 43.7 & 16.0 & 18.3 & 32.8   & 30.7                    \\ \cline{2-8} 
			\multicolumn{1}{|c|}{}                            & SWDA                             & 40.1  & 44.2 & 16.6 & \underline{23.2} & \underline{41.3}   & 33.1                    \\ \cline{2-8} 
			\multicolumn{1}{|c|}{}                            & SCL                              &  33.5  &  48.1 & \underline{18.2} & 15.0 &   28.9   &           28.7         \\ \cline{2-8} 
			\multicolumn{1}{|c|}{}                            & \multicolumn{1}{|c|}{PBDA} & 37.4   & \textbf{54.7}  & 17.2  & 22.5  & 38.5    & \underline{34.1}                     \\ \cline{2-8} 
			\multicolumn{1}{|c|}{}                            & SADA                              & 37.9 & 52.7  & 14.5  & 16.1  &  26.2   &          29.5           \\ \cline{2-8} 
			\multicolumn{1}{|c|}{}                            & \textbf{Ours} & \textbf{47.7}   & \underline{53.4}  & \textbf{19.1}  & \textbf{30.2}  & \textbf{49.3}    & \textbf{39.9}            \\ \hline
	\end{tabular}}
\end{table}

\subsection{Qualitative Results}
\begin{figure}
	\centering
	\begin{subfigure}[b]{.243\textwidth}
		\centering
		% include first image
		\includegraphics[width=\linewidth]{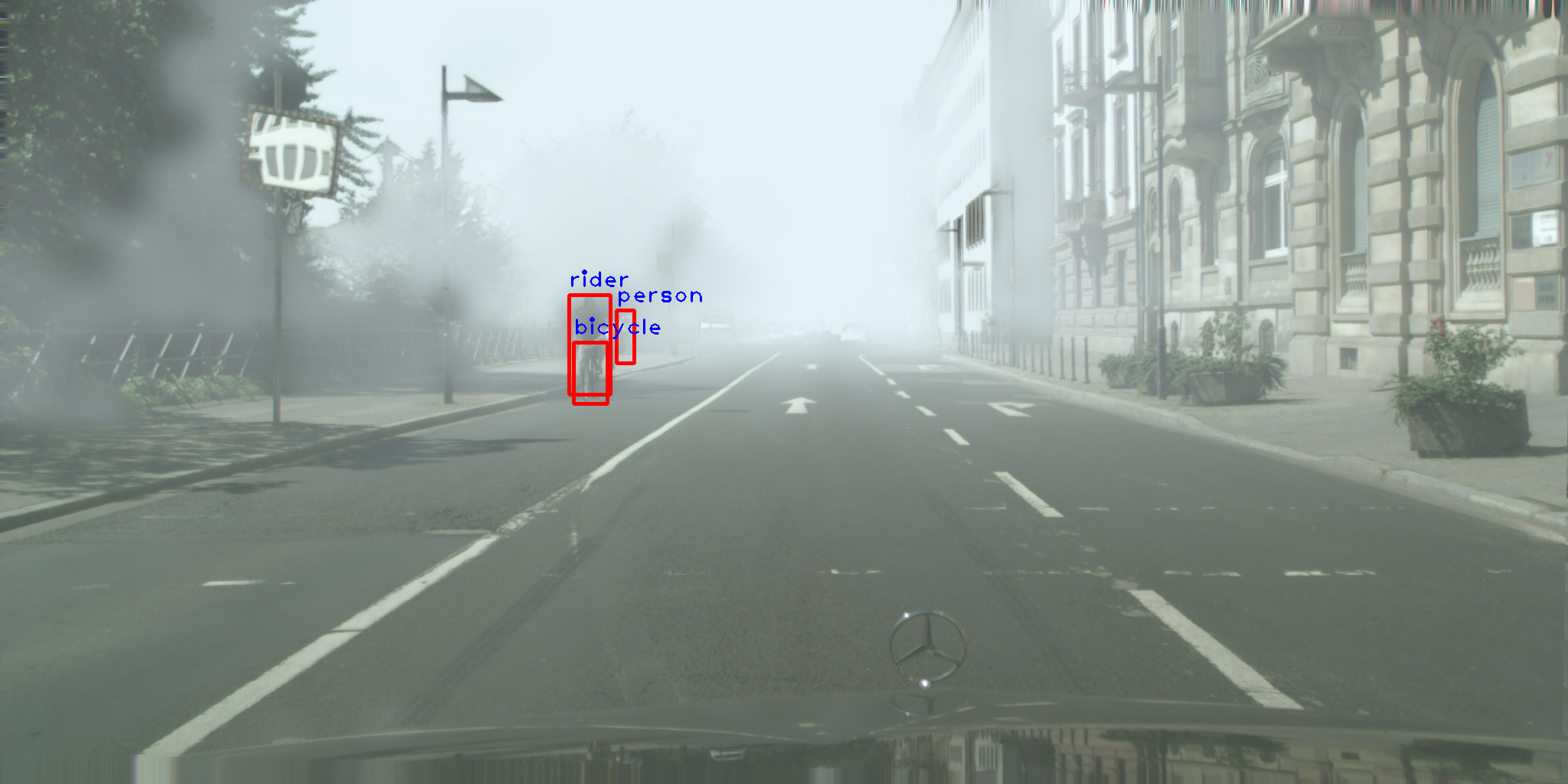}  
		\caption{DA-Faster}
	\end{subfigure}
	\begin{subfigure}[b]{.243\textwidth}
		\centering
		% include first image
		\includegraphics[width=\linewidth]{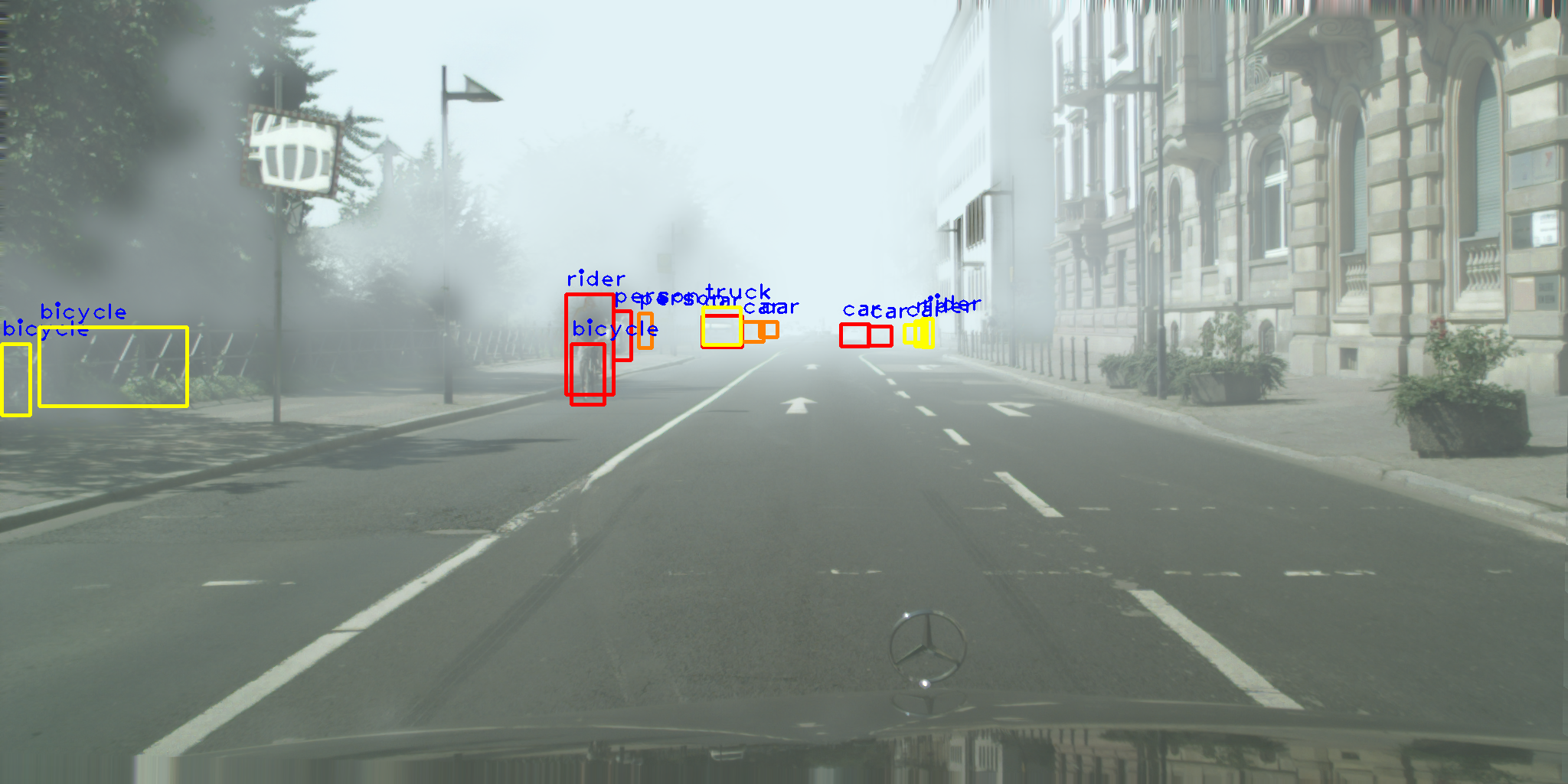} 
		\caption{SADA}
	\end{subfigure}
	\begin{subfigure}[b]{.243\textwidth}
		\centering
		% include first image
		\includegraphics[width=\linewidth]{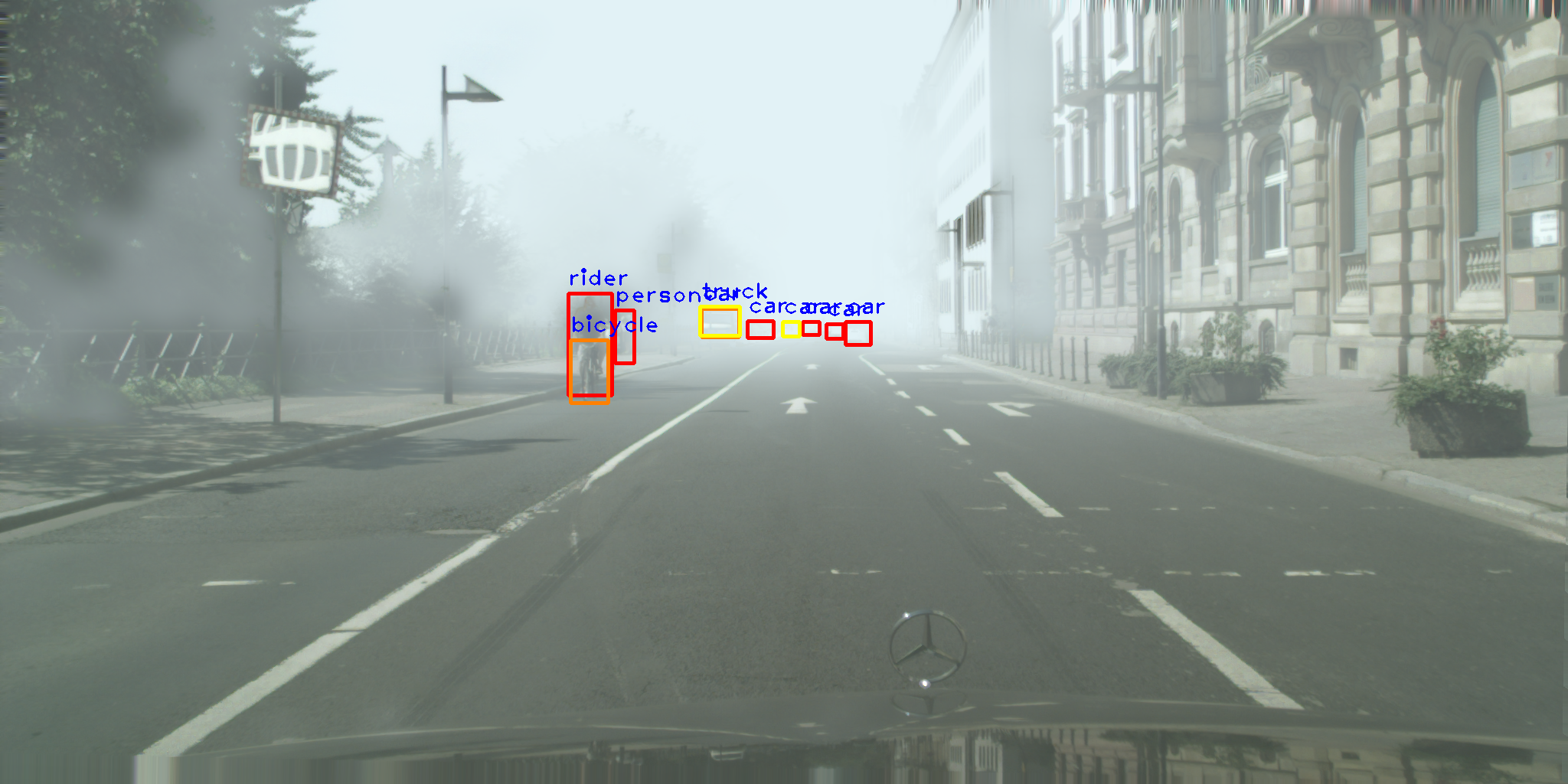}
		\caption{\textbf{Our Result}}
	\end{subfigure}
	\begin{subfigure}[b]{.243\textwidth}
		\centering
		% include second image
		\includegraphics[width=\linewidth]{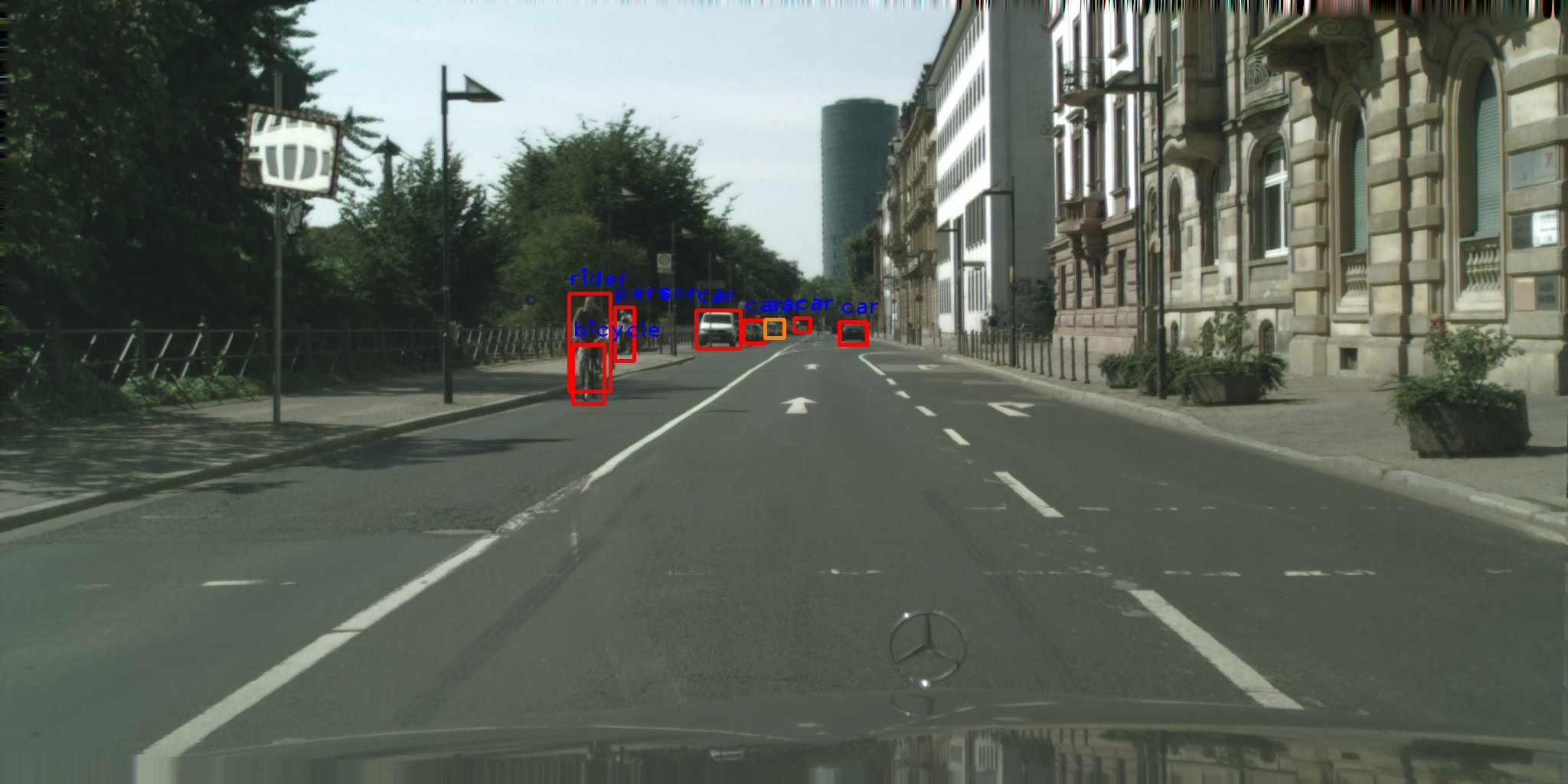} 
		\caption{Upperbound}
	\end{subfigure}
	\\
	\begin{subfigure}[b]{.243\textwidth}
		\centering
		% include third image
		\includegraphics[width=\linewidth]{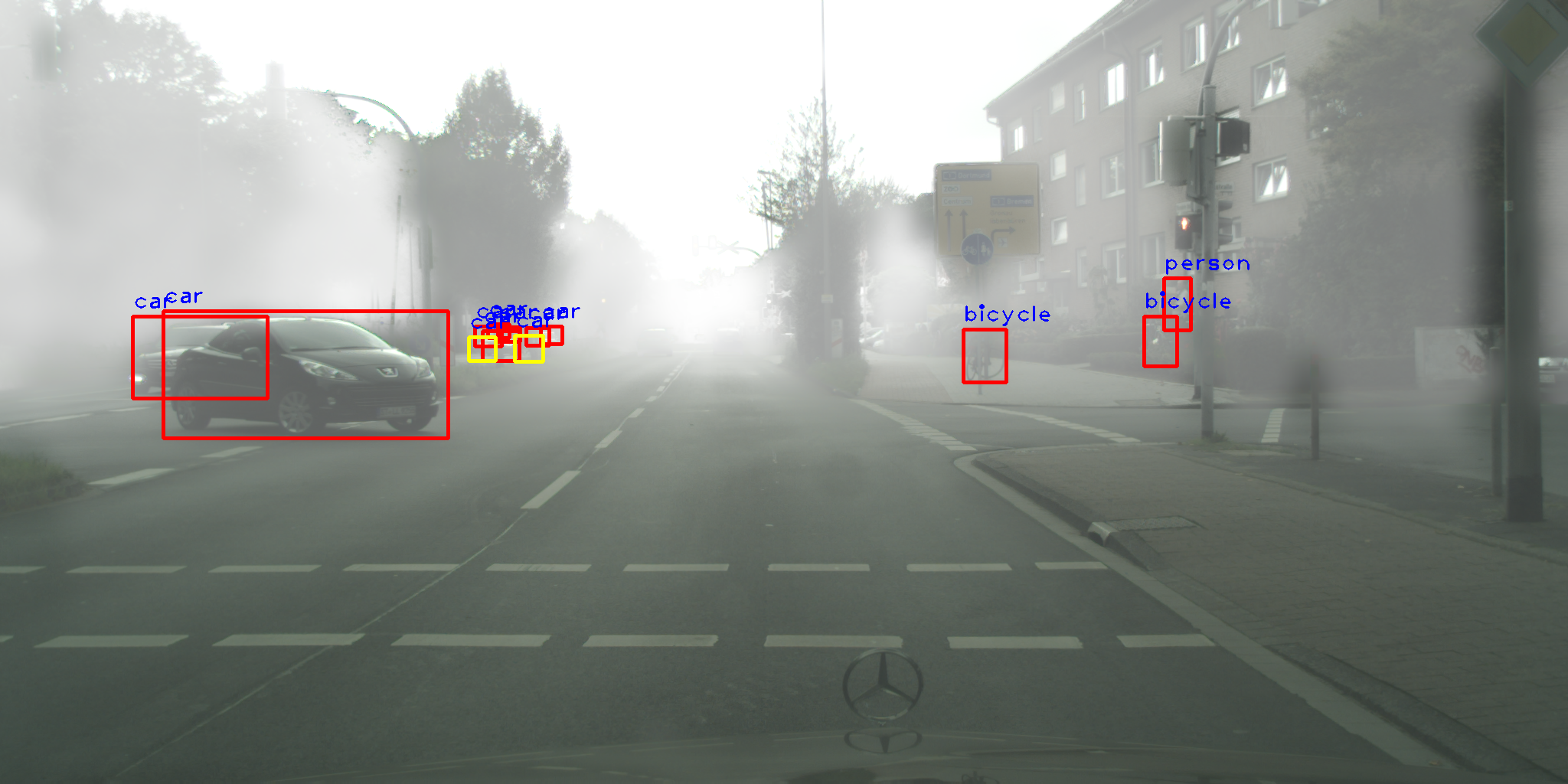}  
		\caption{DA-Faster}
		\label{fig:sub-third}
	\end{subfigure}
	\begin{subfigure}[b]{.243\textwidth}
		\centering
		% include first image
		\includegraphics[width=\linewidth]{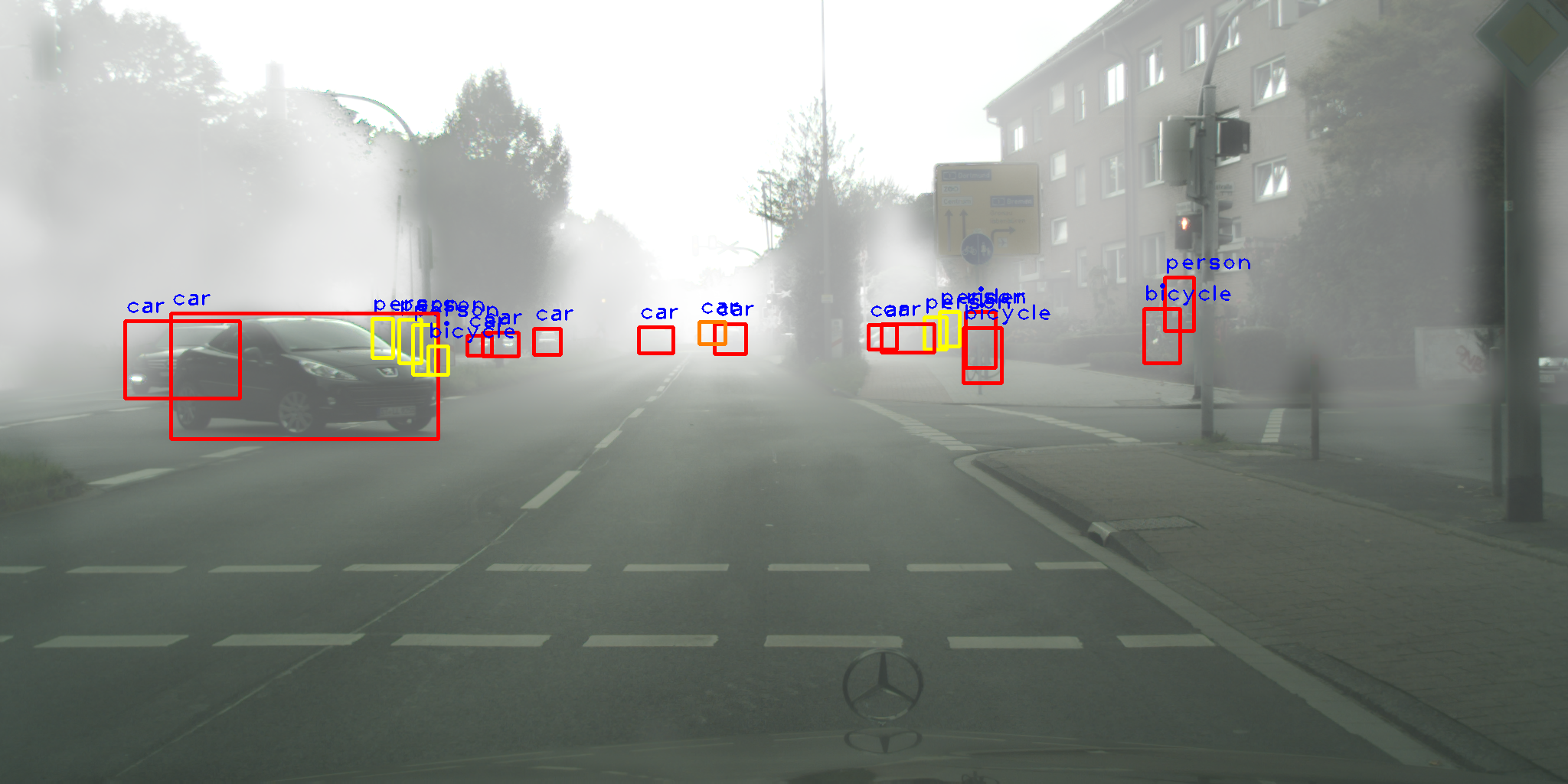}  
		\caption{SADA}
	\end{subfigure}
	\begin{subfigure}[b]{.243\textwidth}
		\centering
		% include first image
		\includegraphics[width=\linewidth]{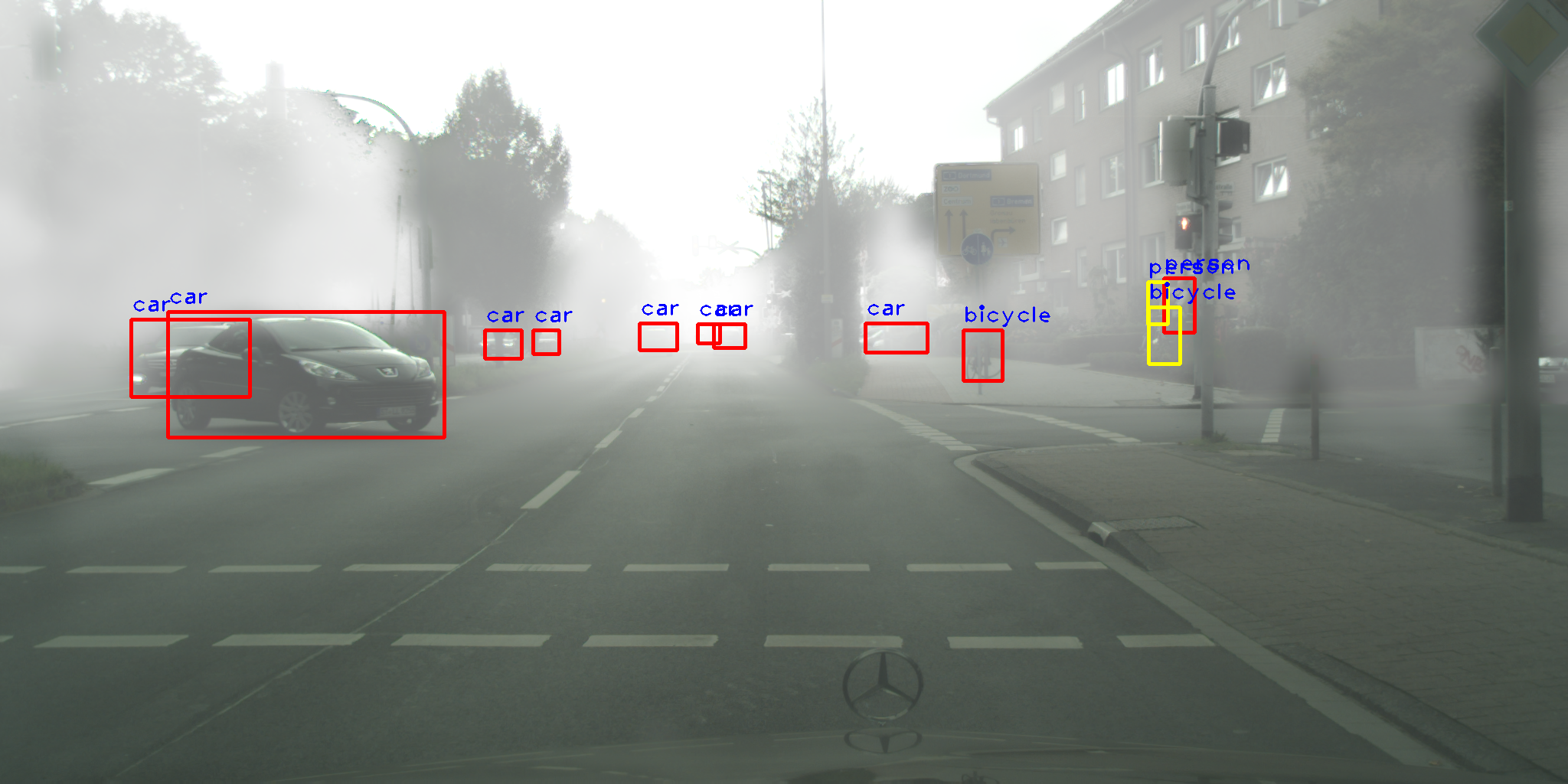}  
		\caption{\textbf{Our Result}}
	\end{subfigure}
	\begin{subfigure}[b]{.243\textwidth}
		\centering
		% include third image
		\includegraphics[width=\linewidth]{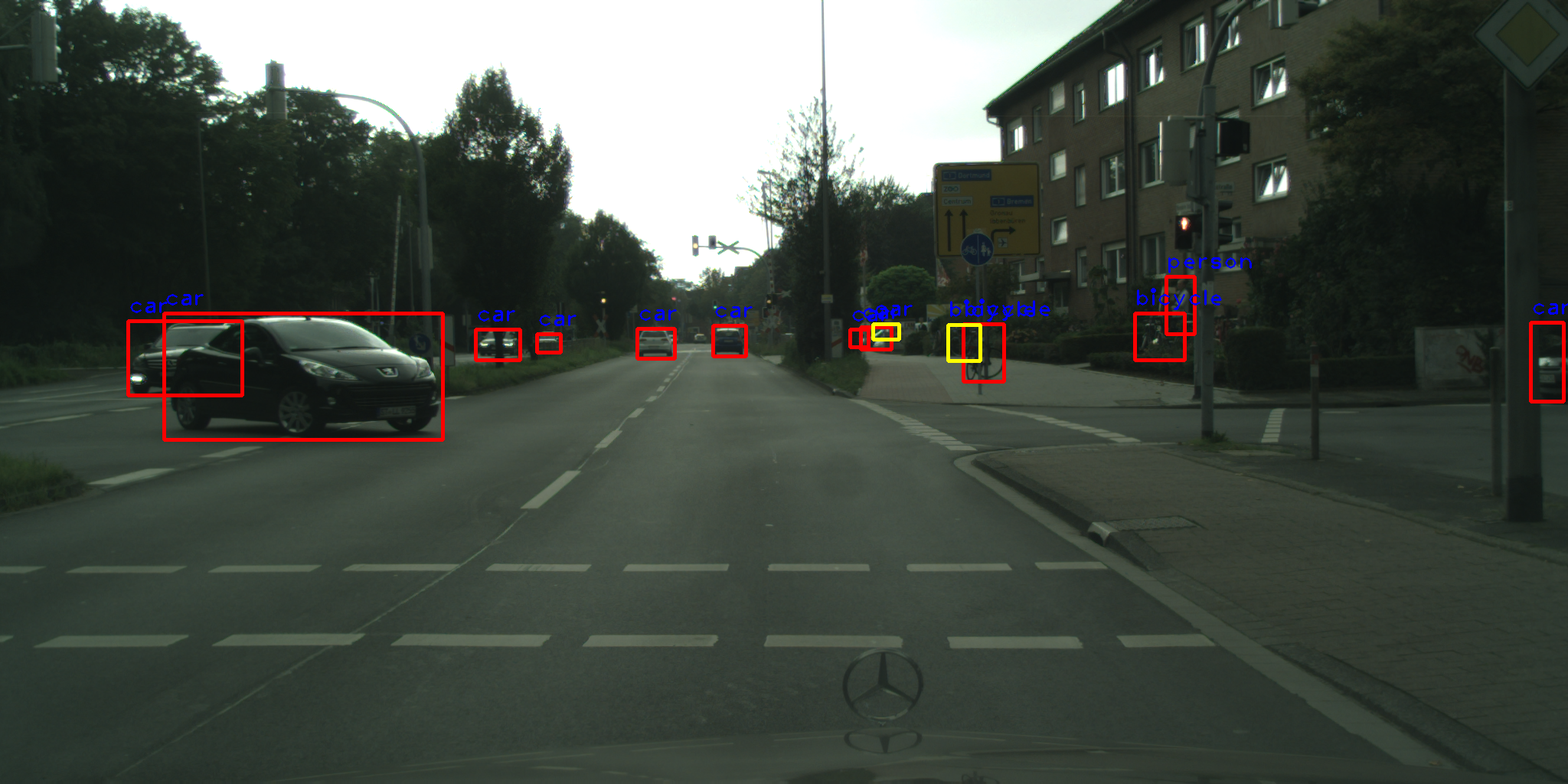}  
		\caption{Upperbound}
	\end{subfigure}
	\\
	\begin{subfigure}[b]{.327\textwidth}
		\centering
		% include third image
		\includegraphics[width=\linewidth,height=2.5cm]{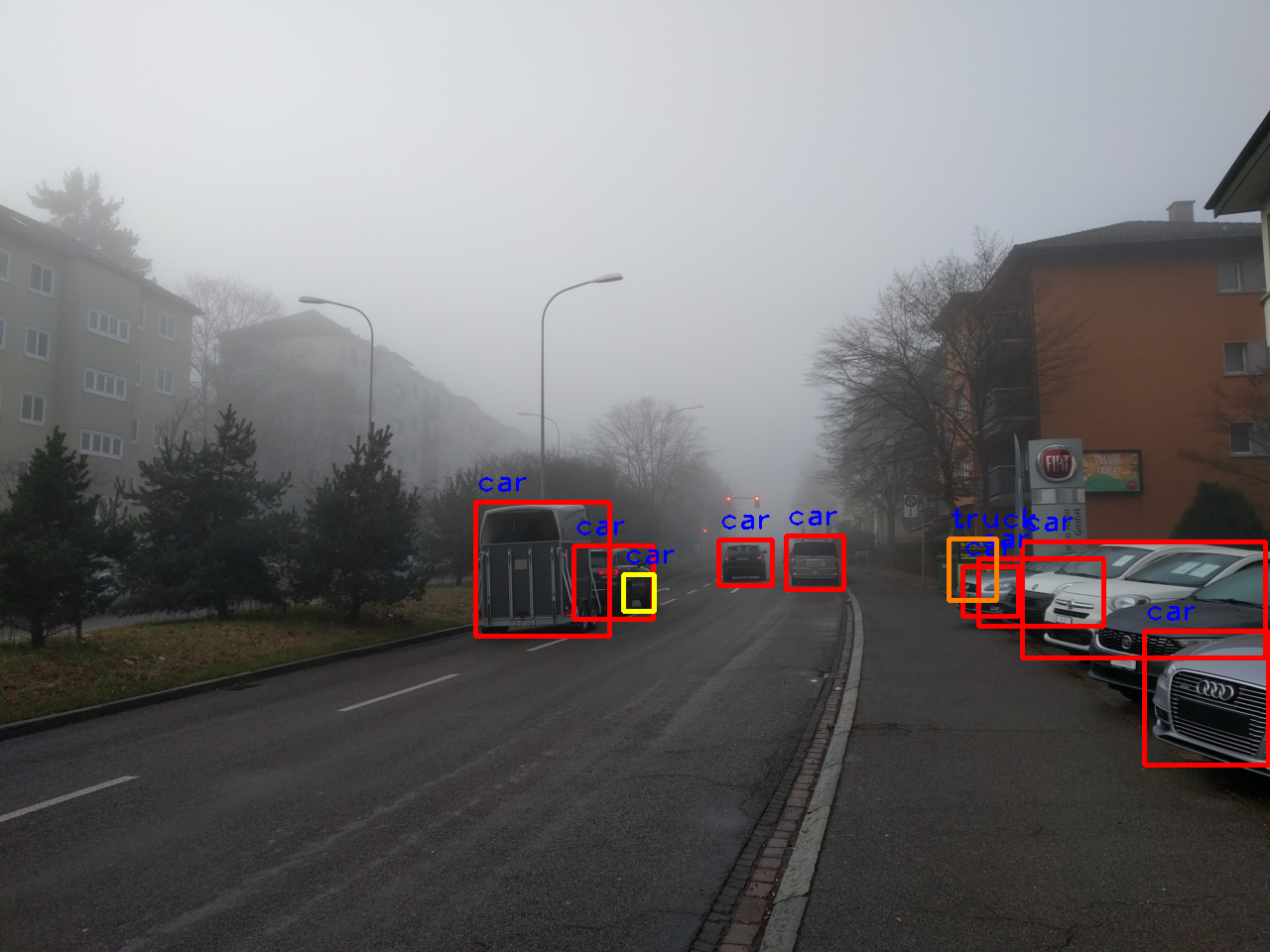}  
		\caption{DA-Faster}
	\end{subfigure}
	\begin{subfigure}[b]{.327\textwidth}
		\centering
		% include first image
		\includegraphics[width=\linewidth,height=2.5cm]{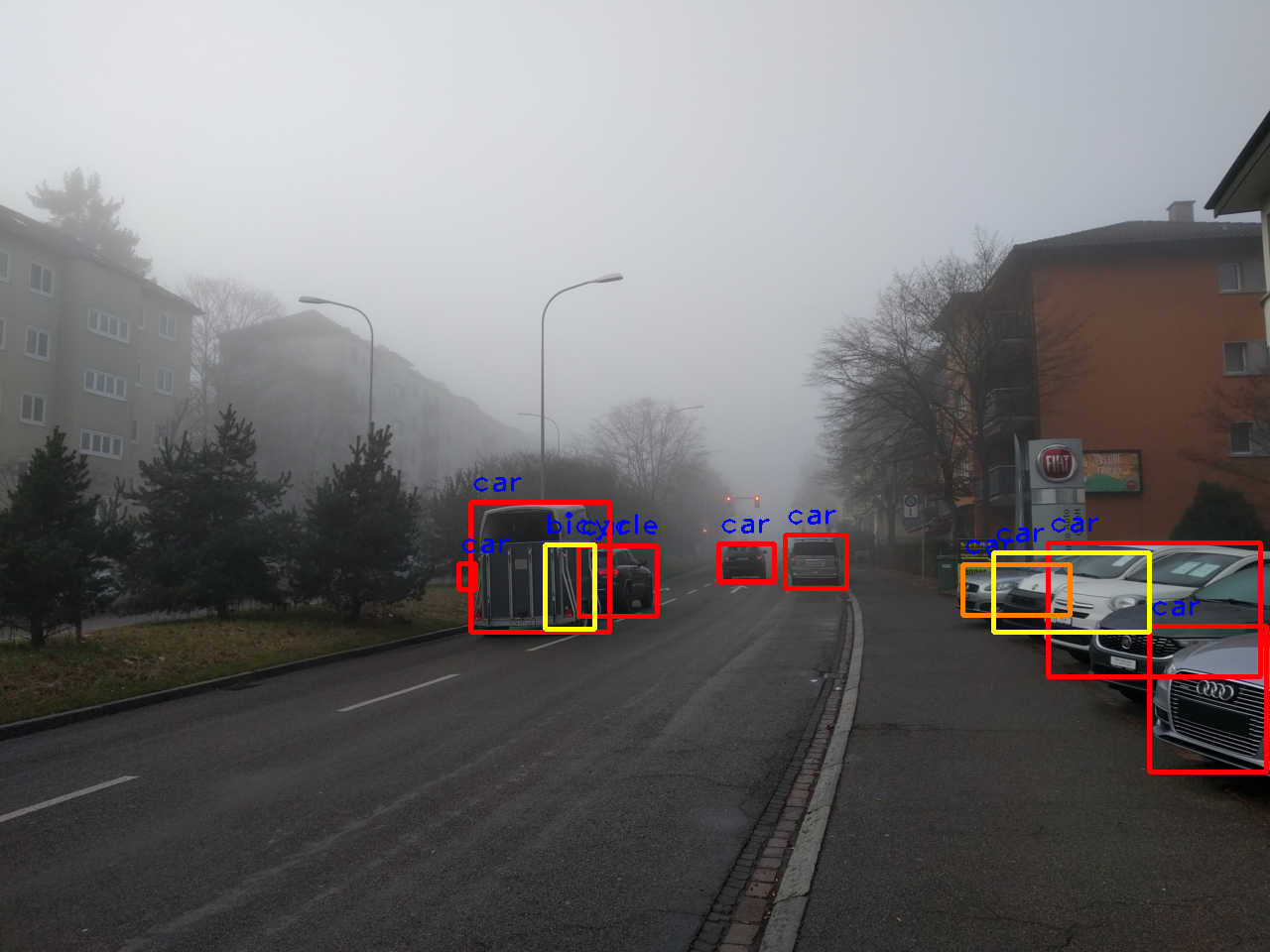}  
		\caption{SADA}
	\end{subfigure}
	\begin{subfigure}[b]{.327\textwidth}
		\centering
		% include first image
		\includegraphics[width=\linewidth,height=2.5cm]{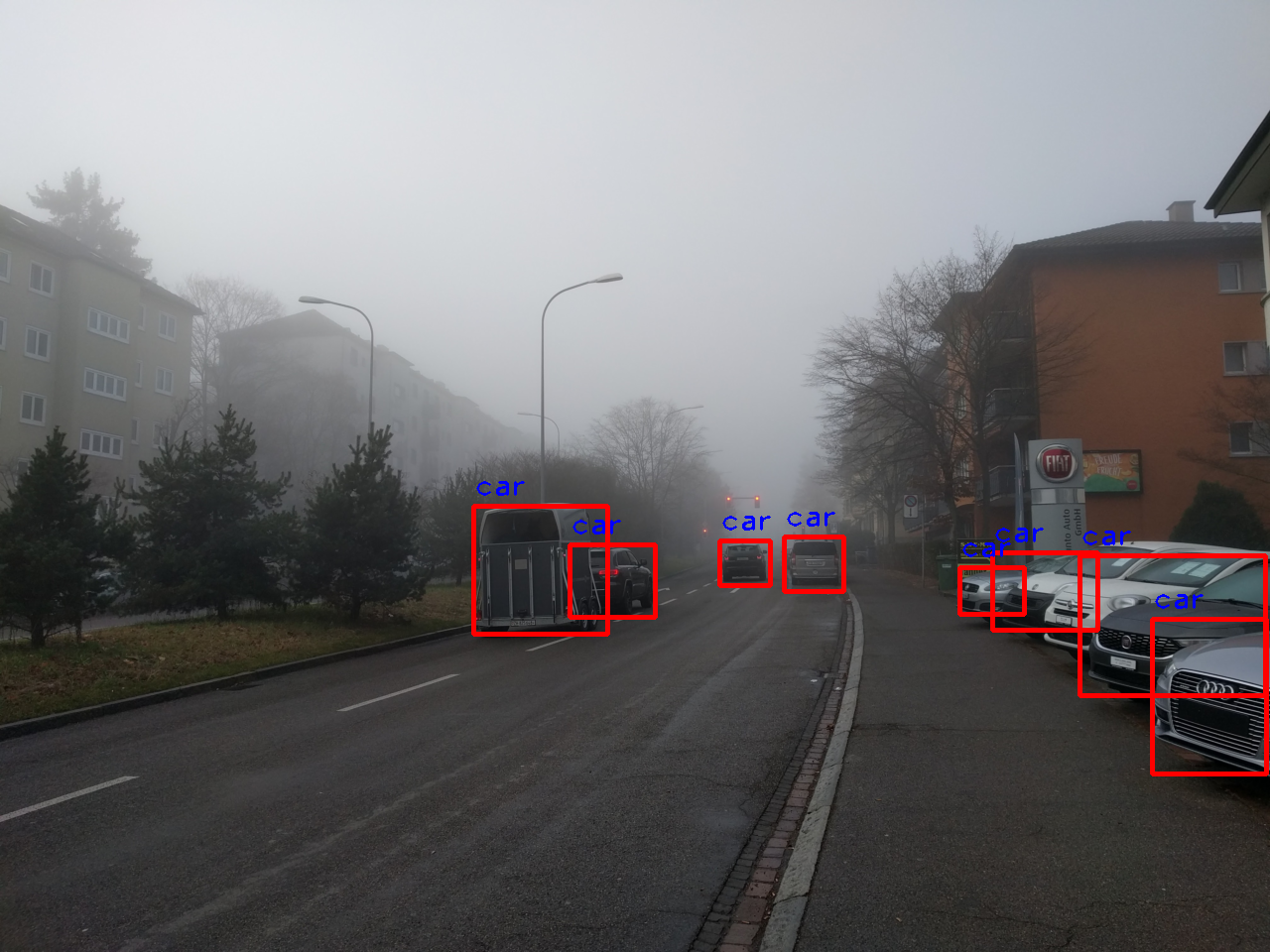}  
		\caption{\textbf{Our Result}}
	\end{subfigure}
	\\
	\begin{subfigure}[b]{.327\textwidth}
		\centering
		% include third image
		\includegraphics[width=\linewidth,height=2.5cm]{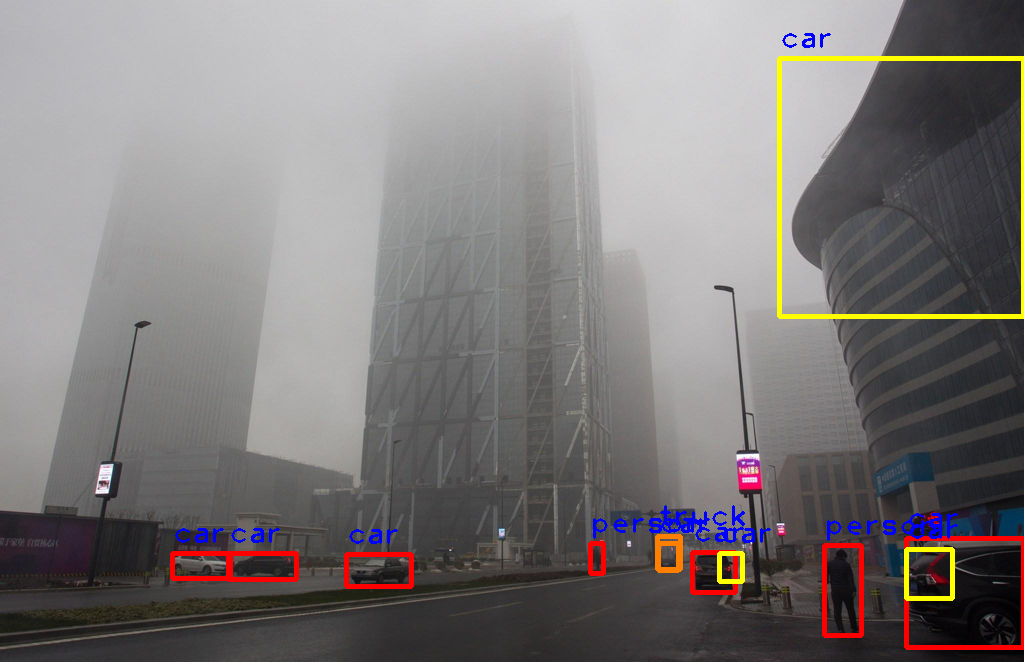}  
		\caption{DA-Faster}
	\end{subfigure}
	\begin{subfigure}[b]{.327\textwidth}
		\centering
		% include first image
		\includegraphics[width=\linewidth,height=2.5cm]{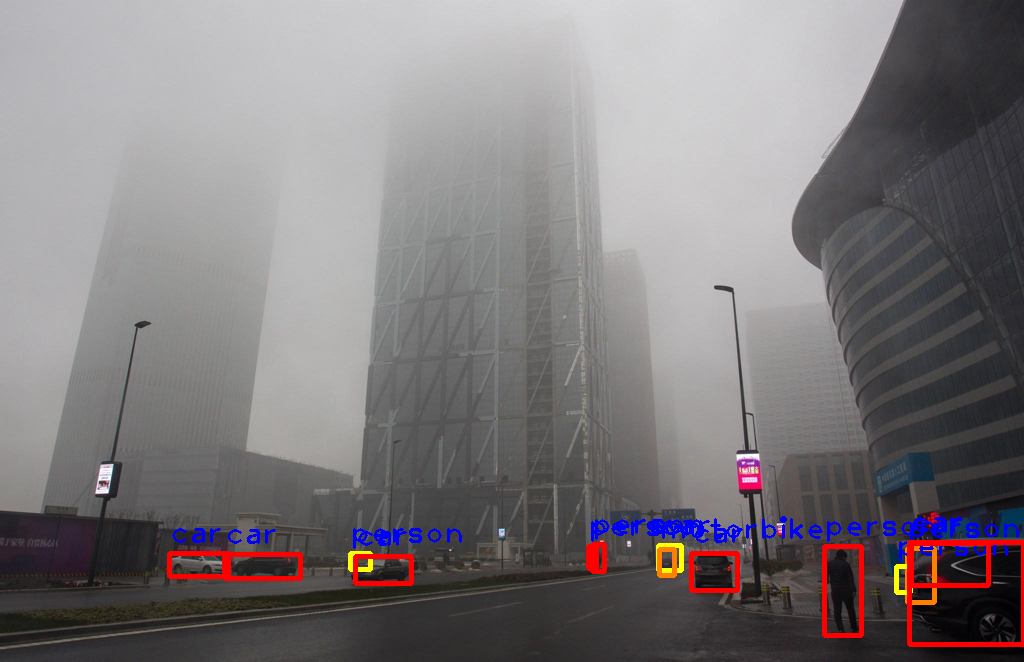}  
		\caption{SADA}
	\end{subfigure}
	\begin{subfigure}[b]{.327\textwidth}
		\centering
		% include first image
		\includegraphics[width=\linewidth,height=2.5cm]{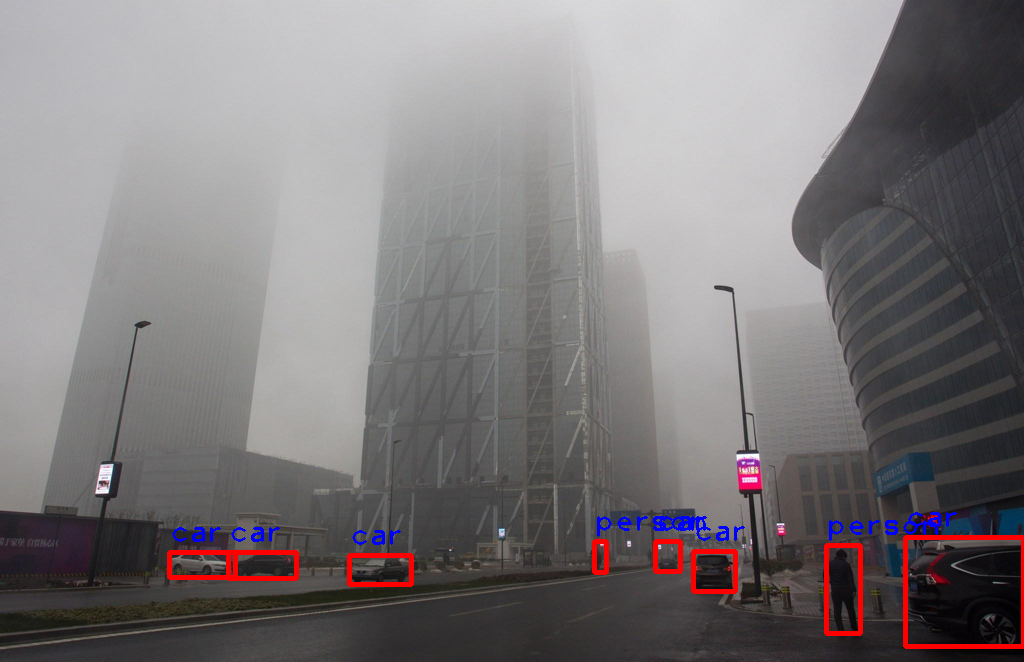}  
		\caption{\textbf{Our Result}}
	\end{subfigure}
	\\
	\begin{subfigure}[b]{.327\textwidth}
		\centering
		% include third image
		\includegraphics[width=\linewidth]{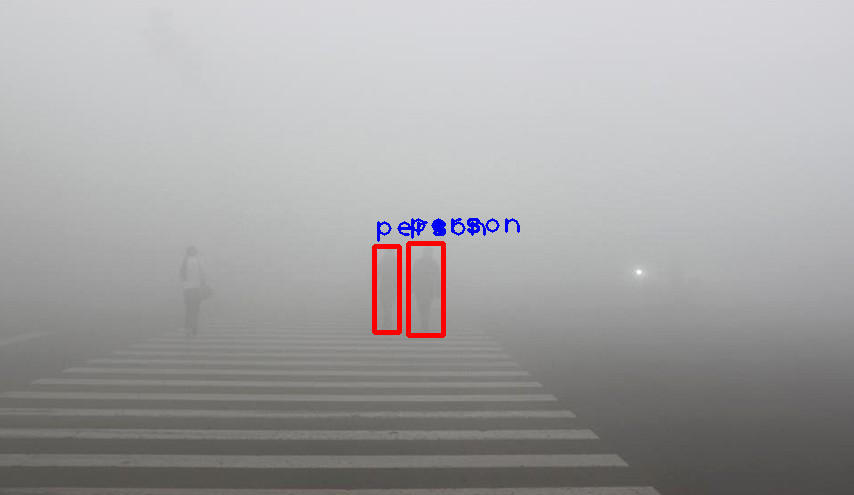}  
		\caption{DA-Faster}
	\end{subfigure}
	\begin{subfigure}[b]{.327\textwidth}
		\centering
		% include first image
		\includegraphics[width=\linewidth]{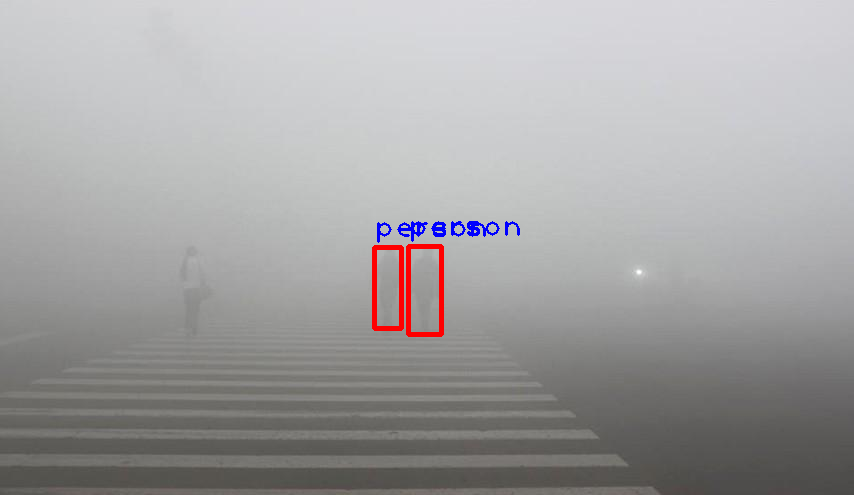}  
		\caption{SADA}
	\end{subfigure}
	\begin{subfigure}[b]{.327\textwidth}
		\centering
		% include first image
		\includegraphics[width=\linewidth]{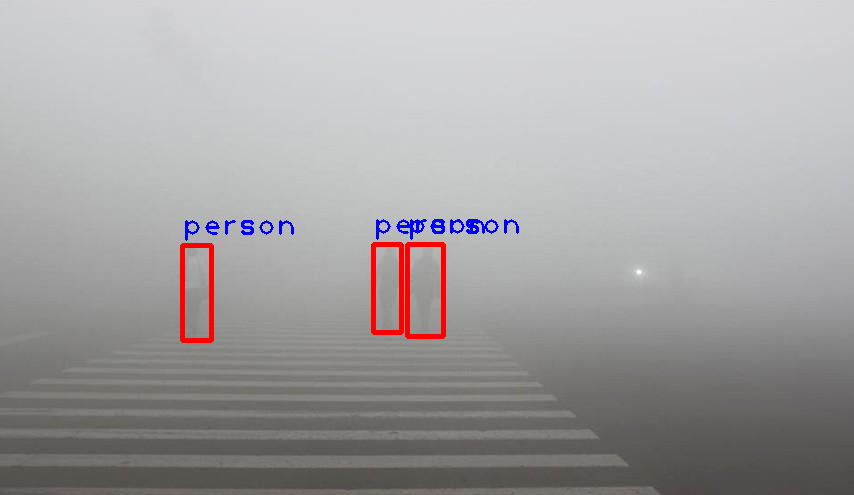}  
		\caption{\textbf{Our Result}}
	\end{subfigure}
	\\
	\begin{subfigure}[b]{.3\textwidth}
		\centering
		% include third image
		\includegraphics[width=\linewidth]{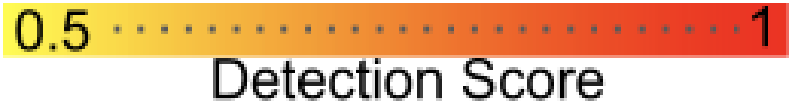} 
	\end{subfigure}
	\caption{\label{fig:qualiresults} Comparisons with DA-Faster\cite{chen2018domain}, SADA\cite{chen2021scale}, and Upperbound. The first two rows are the comparison on Cityscapes $\rightarrow$ Foggy Cityscapes. Our model detects more objects and reduces false positive predictions and approaches the Upperbound performance. The last three rows are the comparison on real world images. Our model has more true positive detections and less false positive detections. Different bounding box's colour represents a different confidence score.}
\end{figure}

The qualitative results are presented in \cref{fig:qualiresults}.
We evaluate our model on both synthetic and real world datasets, and compare it with DA-Faster\cite{chen2018domain} and SADA\cite{chen2021scale}.
For the synthetic dataset, we also compare the model with Upperbound to visualize how close their predictions are.
Note again that, Upperbound is the Faster-RCNN model trained on the clear training set and tested on the clear testing set.
We can see that our method can detect more objects compared to DA-Faster.
Both SADA and our method can detect most of the object instances in fog.
However, we can see that SADA generated some false predictions.
Our method removed some false predictions, and thus the final object detection performance is approaching Upperbound. 

\begin{figure}[ht]
	\centering
	\begin{subfigure}{0.495\linewidth}
		\includegraphics[width=1\linewidth,height=3.4cm]{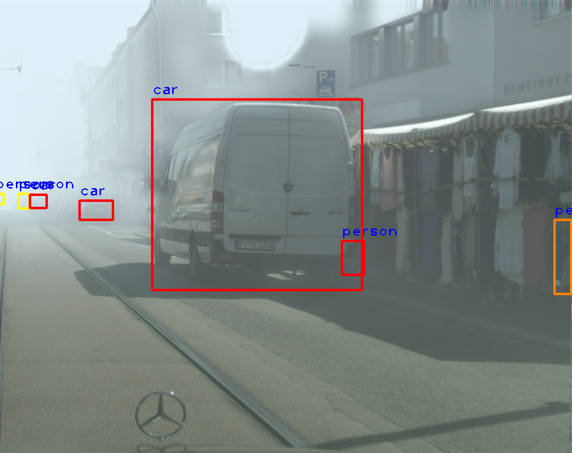}
		\caption{Without $\mathcal{L}_{\text{rec}}$}
	\end{subfigure}
	\begin{subfigure}{0.495\linewidth}
		\includegraphics[width=1\linewidth,height=3.4cm]{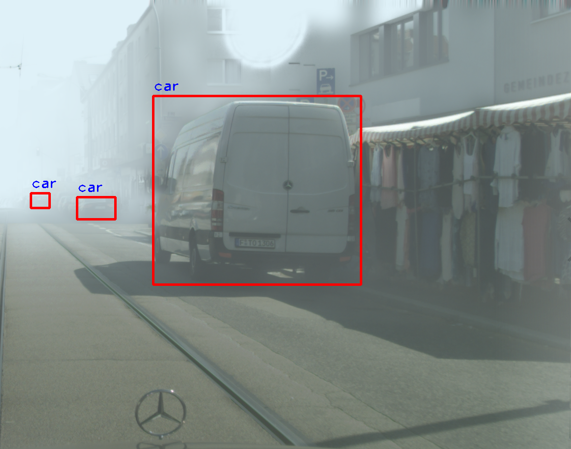}
		\caption{With $\mathcal{L}_{\text{rec}}$}
	\end{subfigure}
	\\
	\begin{subfigure}{0.495\linewidth}
		\includegraphics[width=1\linewidth,height=3cm]{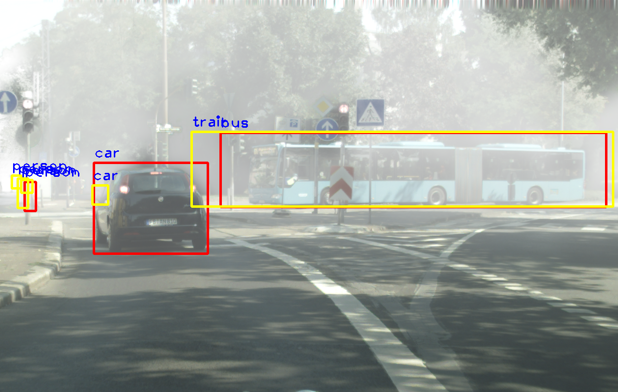}
		\caption{Without $\mathcal{L}_{\text{rec}}$}
	\end{subfigure}
	\begin{subfigure}{0.495\linewidth}
		\includegraphics[width=1\linewidth,height=3cm]{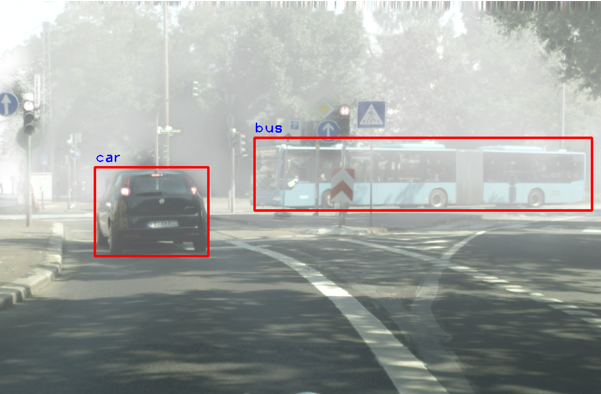}
		\caption{With $\mathcal{L}_{\text{rec}}$}
	\end{subfigure}
	\\
	\begin{subfigure}[b]{.3\textwidth}
		\centering
		% include third image
		\includegraphics[width=\linewidth]{score.PNG} 
	\end{subfigure}
	\caption{\label{fig:failedfog} Ablation studies on our reconstruction loss. (a)(c): False positive predictions when there is no reconstruction, from model (DA+DEB+Consist). (b)(d): After we added our reconstruction loss. The false positive predictions are reduced.}
\end{figure}

\begin{table}
	\centering
	\caption{\label{tab:abla1}Ablation study of our model against the Foggy Cityscapes testing set. mAP is used to analyze the effectiveness of each loss.}
	\resizebox{0.5\textwidth}{!}{\begin{tabular}{|c|c|c|c|c|c|}
			\hline
			DA & DEB & Consist &  Reconst & PL &  mAP   \\ \hline
			
			\checkmark    &     &         &              &          &         42.6  \\ \hline
			\checkmark  &  \checkmark   &         &              &          &         43.3 \\ \hline
			\checkmark     &  \checkmark   &   \checkmark      &              &       &            45.3 \\ \hline
			\checkmark     &   \checkmark  &   \checkmark      &       \checkmark        &     &             45.8 \\ \hline
			\checkmark     &   \checkmark  &   \checkmark      &       \checkmark        &     \checkmark    &          47.6 \\  \hline
	\end{tabular}}
\end{table}

\subsection{Ablation Studies} 
\cref{tab:abla1} shows the  ablation studies on Foggy Cityscapes to demonstrate the importance of each loss. The check mark indicates which losses are involved.
In the table, DA represents the performance with domain discriminator only, DEB represents the depth recovery loss, Consist represents the transmission-depth consistency loss.
Reconst represents the reconstruction loss using the pseudo ground-truth images generated from DCP \cite{he2010single}.
We also provide \cref{fig:failedfog} to show the comparisons between the models with and without our reconstruction module.
PL means we include target predictions as pseudo labels for training.
As one can notice, each of the losses improves the overall mAP.
The network has a performance gain with only pseudo ground-truths of transmission maps, depth maps, and defogged images.
This once again shows that our method works without the need of precise ground-truths.
If we use the ground-truths of  transmission maps, depth maps, and defogged images (which are actually available for Foggy Cityscape), our performance reaches overall 49.2 mAP.
The weights of the losses are chosen to ensure that all the losses will contribute to the training properly. If we set $a$ from 10 to 1, or we set $b$ from 1 to 0.1, the performance drops by around 1 mAP. If we set $c$ from 1 to 0.1, the performance drops by 2 mAP. Setting $\lambda$ to be 0.1 is recommended by a few DA papers. The performance drops below 40 mAP if $\lambda$ becomes too large. In our method, the weights are set empirically.

%------------------------------------------------------------------------- 

\section{Conclusion}
\label{sec:conclusion}
We have proposed a novel DA method with a reconstruction as a regularization, to develop an object detection network which is robust for fog or haze conditions.
To address the problem that DA process can suppress depth information, we proposed the DEB to recover it.
We proposed the transmission-depth consistency loss to reinforce the transmission map based DA to follow the target image's depth distribution.
We integrated a reconstruction module to our DA backbone to reconstruct a clear image of the target image and reduce the false object instance features.
We involved target domain knowledge into DA, by reusing reliable target predictions and enforcing consistent detection.
We evaluated the framework on several benchmark datasets showing that our method outperforms the state-of-the-art DA methods.

%
% ---- Bibliography ----
%
% BibTeX users should specify bibliography style 'splncs04'.
% References will then be sorted and formatted in the correct style.
%
\bibliographystyle{splncs04}
\bibliography{egbib}
\end{document}